\documentclass{article}

\usepackage[nonatbib, preprint]{neurips_2025}

\usepackage[utf8]{inputenc} 
\usepackage[T1]{fontenc}    
\usepackage{hyperref}       
\usepackage{url}            
\hypersetup{colorlinks=true, linkcolor=red, anchorcolor=blue, citecolor=blue}
\usepackage{booktabs}       
\usepackage{amsfonts}       
\usepackage{nicefrac}       
\usepackage{microtype}      
\usepackage{pifont}
\usepackage{graphicx}
\usepackage{amsmath}
\usepackage{multirow}
\usepackage{makecell}
\usepackage{pifont}
\usepackage{placeins}
\usepackage{tikz}
\usepackage{colortbl} 
\usepackage{nicefrac}
\usepackage{dsfont}
\usepackage{marvosym}
\usepackage{fdsymbol}
\usepackage{nicematrix}
\usepackage{bbm}
\usepackage[capitalize]{cleveref}
\usepackage{enumitem}

\usepackage{xspace}
\newcommand{\sqbullet}{\ding{113}} 

\newcommand{\srcc}[1]{/\hspace{1pt}#1}

\definecolor{myIDBcolor}{RGB}{245,248,255}
\newcommand{\best}[1]{\textbf{\textcolor{red}{#1}}}
\newcommand{\secondbest}[1]{\underline{\textcolor{blue}{#1}}}
\newcommand{\ours}{\textbf{Q-Tacit}\xspace}

\newcommand{\LVRtag}{\colorbox{orange!15}{\texttt{<lvr>}}}
\newcommand{\LVRstart}{\colorbox{orange!15}{\texttt{<|lvr\_start|>}}}
\newcommand{\LVRend}{\colorbox{orange!15}{\texttt{<|lvr\_end|>}}}
\newcommand{\answertagl}{\texttt{<answer>}}
\newcommand{\answertagr}{\texttt{</answer>}}

\title{Q-Tacit: Image Quality Assessment \\via Latent Visual Reasoning}

\author{Yuxuan Jiang\textsuperscript{1,\dag},
Yixuan Li\textsuperscript{1,\dag,~\Letter}, Hanwei Zhu\textsuperscript{2}, 
Siyue Teng\textsuperscript{1}, Fan Zhang\textsuperscript{1}, David Bull\textsuperscript{1}\\
\textsuperscript{1} Visual Information Lab, University of Bristol\\
\textsuperscript{2} Nanyang Technological University\\
{\tt\small\{yuxuan.jiang, siyue.teng, fan.zhang, dave.bull\}@bristol.ac.uk}\\
{\tt\small yixuanli423@gmail.com}\\
{\tt\small hanwei.zhu@ntu.edu.sg}\\
\url{https://github.com/YuxuanJJ/Q-Tacit}
}

\begin{document}
\maketitle
\footnotetext[1]{\dag Equal contribution.}
\footnotetext[2]{\Letter: Corresponding authors}

\begin{abstract}
Vision-Language Model (VLM)-based image quality assessment (IQA) has been significantly advanced by incorporating Chain-of-Thought (CoT) reasoning. Recent work has refined image quality reasoning by applying reinforcement learning (RL) and leveraging active visual tools. However, such strategies are typically language-centric, with visual information being treated as static preconditions. Quality-related visual cues often cannot be abstracted into text in extenso due to the gap between discrete textual tokens and quality perception space, which in turn restricts the reasoning effectiveness for visually intensive IQA tasks. In this paper, we revisit this by asking the question, \textit{``Is natural language the ideal space for quality reasoning?''} and, as a consequence, we propose \ours, a new paradigm that elicits VLMs to reason beyond natural language in the latent quality space. Our approach follows a synergistic two-stage process: (i) injecting structural visual quality priors into the latent space, and (ii) calibrating latent reasoning trajectories to improve quality assessment ability. Extensive experiments demonstrate that \ours can effectively perform quality reasoning with significantly fewer tokens than previous reasoning-based methods, while achieving strong overall performance. This paper validates the proposition that language is not the only compact representation suitable for visual quality, opening possibilities for further exploration of effective latent reasoning paradigms for IQA. Source code will be released to support future research.

\end{abstract}


\section{Introduction}
\label{sec:intro}

Image quality assessment (IQA) plays an essential role in modern computer vision ~\cite{li2025qinsight,wu2025visualqualityr1,wu2023QAlign,zhuadaptive,tian2025aigivc}, attempting to establish a quantitative mapping between a digital image and its human subjective evaluation. Echoing Moravec's paradox~\cite{moravec1988mindchildren}, \textit{it is comparatively easy to make computers exhibit adult-level performance on intelligence tests, while difficult to give them the perception of a one-year-old}, IQA is instinctive to humans, yet requires complex reasoning to create accurate objective measures. 

No-Reference (NR) IQA, which assesses images without a reference, has evolved for decades from knowledge-driven~\cite{mittal2012no,mittal2012making} to data-driven approaches~\cite{talebi2018nima,ke2021musiq}, tackling generalization~\cite{zhu2020metaiqa,chen2021no,chen2025toward}, active fine-tuning~\cite{wang2021troubleshooting}, and continual adaptation~\cite{zhang2022continual} problems. Despite these advances, such methods remain easily challenged by out-of-distribution (OOD) scenarios. Their inference primarily depends on the implicit function of the learned representation, which can lack delicate reasoning\footnote{In line with the prevalent usage in VLM research, ``reasoning'' here denotes an inspectable explicit inference trace that links vision to model outputs.} for more reliable quality inference.

\begin{figure*}[!t]
  \centering
  \includegraphics[width=\linewidth]{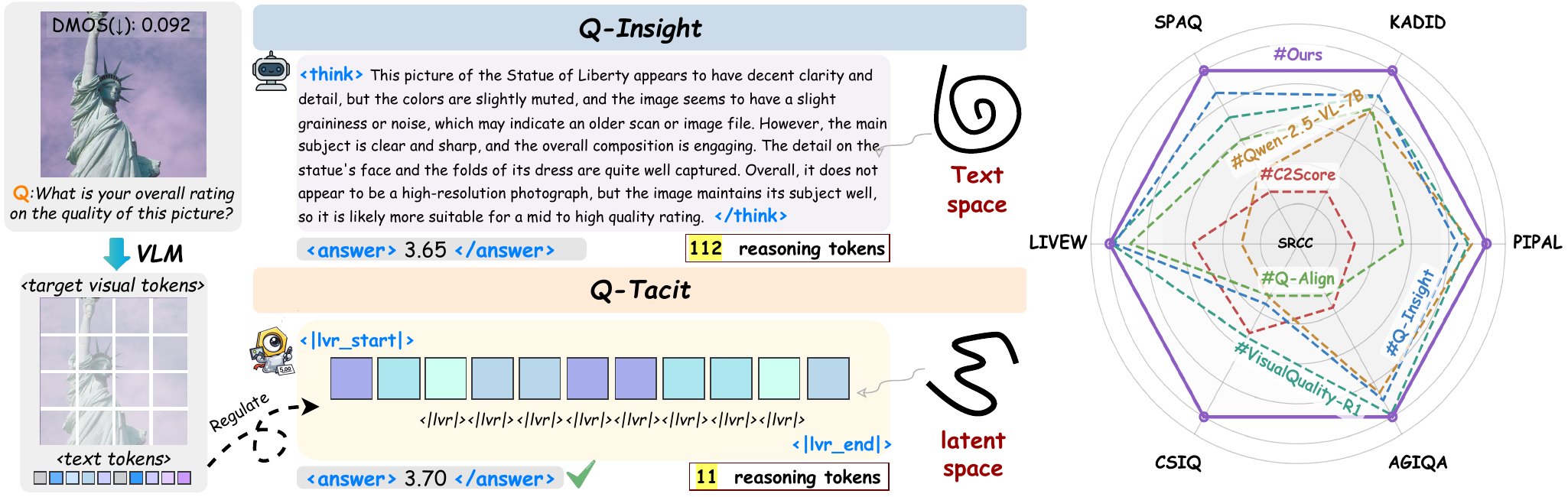}
  \caption{\textbf{Motivation and overview of Q-Tacit.} We compare \ours and Q-Insight~\cite{li2025qinsight}, which perform quality reasoning in a latent space and a text space, respectively. \texttt{<|lvr\_start|>} and \texttt{<|lvr\_end|>} encapsulate the latent reasoning process, and \texttt{<|lvr|>} is a placeholder for a latent slot token that allocates one latent reasoning step. \ours excels at image quality scoring on out-of-distribution datasets (Right) while requiring only 10\% visible token counts compared to Q-Insight (Left).}
  \label{fig:qtacit-overview}
\end{figure*}

Building upon the rapid advancement of large Visual Language Models (VLMs), NR-IQA has leveraged their extensive visual reasoning abilities and cross-modal semantic contextualization for quality evaluation~\cite{sun2023aligning,yu2024rlaif,zhang2025r1vl,zhao2025beyond,bai2025qwen2}. Pre-existing VLM-based methods generally fall into two categories: methods without explicit reasoning, such as Q-Align~\cite{wu2023QAlign}, Compare2Score~\cite{zhuadaptive} and DeQA-Score~\cite{you2025teaching}, and reasoning-induced methods, represented by VisualQuality-R1~\cite{wu2025visualqualityr1}, Q-Insight~\cite{li2025qinsight}, and Q-ponder~\cite{cai2025q}. Amongst these categories, reasoning-induced methods have recently attracted increased attention. Compared to direct inference, these methods usually introduce intermediate rationales as the reasoning route to quality scores. As shown in~\cref{fig:qtacit-overview}, when evaluating an input image, Q-Insight first describes quality-relevant visual cues in natural language, providing additional backing for subsequent quality scoring. In \cite{zhao2025reasoning}, it is demonstrated that aligning images to the Reinforcement Learning (RL)-derived quality-description text space can preserve generalization without reasoning at inference, further suggesting that the improved reasoning signal can accordingly boost quality prediction. Therefore, if quality-awareness can be enhanced by refining the reasoning process, the explicit form of the intermediate process becomes pivotal. Although text-centric reasoning provides certain explainability for IQA, it can be difficult for these methods to faithfully form quality abstractions that go beyond what can be easily verbalized~\cite{li2025latent,li2025latentimplicitvisualreasoning,wang2025monet}. This naturally leads to the following question: \textit{is natural language the ideal space for visual quality reasoning}?

In this paper, we explore the potential of extending VLM quality reasoning beyond text towards the use of visual tokens that directly encode quality information. Rather than improving a model’s ability to verbalize image quality in natural language, we instead drive it to reason in a latent quality space shared by continuous visual tokens and discrete text tokens~\cite{li2025latent,qin2025covt}. A related line of research seeks to enforce stronger visual grounding in quality-assessment reasoning via active tools, such as zooming in~\cite{liang2026zoomiqa}. Although such tools can partially compensate for the loss of visual quality details caused by VLMs' cross-modality bias and interference~\cite{zhang2025beyond,cai2025diagnosing}, they still revert to refining text generation, and thus the fundamental gap persists between images and textual trajectories. Hence, we adopt latent visual reasoning (LVR)~\cite{li2025latent}, a multimodal reasoning scheme that moves reasoning towards a latent quality space, encoding visual semantics relevant to both the input image and the query. LVR has been effective in tasks such as object detection~\cite{wang2025monet} and image generation~\cite{chen2026show}. By exploring reasoning interfaces that couple more tightly with perceptual evidence, VLMs can benefit in three main ways: (i) improved generalization via visually grounded latent reasoning space; (ii) no reliance on dense textual supervision; and (iii) efficient reasoning that requires a significantly lower token budget.

Specifically, we propose \ours to steer VLMs to reason about image quality in the latent quality space. We instantiate a dedicated reasoning space by inserting a sequence of latent tokens, and formulate it as latent-space visual reconstruction. Instead of using visual embeddings as static conditioning, the model performs a series of auto-regressive hidden-state transitions over the latent tokens to reason with query-relevant visual tokens before outputting a final prediction, as shown in~\cref{fig:qtacit-overview}. Such transitions can be regarded as continuous \textit{latent visual thoughts} towards quality, similar to Chain-of-Thought (CoT) in textual space~\cite{li2025latent,li2025latentimplicitvisualreasoning,wang2025monet}. Finally, we calibrate the reasoning trajectories in this latent space, which enables the model to operate directly on continuous perceptual representation, thereby preserving subtle cues to reason out the quality score. Practically, we realize this scheme by introducing a synergistic two-stage process elaborated as follows:

\noindent{\textit{Stage I: Latent Prior Injection.}} We form a latent quality space and teach the model basic latent reasoning patterns through Supervised Fine-Tuning (SFT). We propose to use the Regions of Interest (ROI)-based quality data to supervise the last hidden states and force latent tokens to reconstruct quality-relevant visual evidence, thereby linking latent reconstruction and perceptual scoring.

\noindent{\textit{Stage II: Quality-aligned Calibration.}} We then perform the latent Group Relative Policy Optimization (GRPO) algorithm~\cite{shao2024deepseekmath,guo2025deepseek} to calibrate latent reasoning trajectories towards human quality perception. We design a gaussian-based dense quality scoring reward and combine it with a format-compliance reward to jointly optimize quality reasoning without requiring dense manual annotations. \\
As shown in \cref{fig:qtacit-overview}, our Q-Tacit exhibits remarkable quality prediction performance on OOD datasets while demanding significantly reduced reasoning tokens. Our empirical results further reveal that the latent space compactly encodes quality-relevant evidence by tightly coupling reasoning embeddings with visual embeddings, leading to a reasoning space that is more sensitive to subtle local degradations. Our contributions are three-fold:

\begin{itemize}[label=\sqbullet, leftmargin=*, itemsep=0pt, topsep=2pt]
\item We propose \ours, the first reasoning-induced IQA framework in the latent space. Distinguished from previous methods which conduct text-centric quality reasoning, our approach achieves superior quality prediction performance in the latent visual space with incremental reasoning token saving.

\item We introduce a continuous latent space framed by latent evidence reconstruction and aggregation, which embeds quality-relevant perception and general visual understanding simultaneously. This enables effective and scalable image quality reasoning towards human perception.  

\item We conduct extensive experiments to validate the effectiveness of reasoning quality in the latent space. Q-Tacit exhibits consistent improvements on quality prediction compared with previous methods across diversified distortions.

\end{itemize}

\section{Related Work}
\label{sec:related_work}

\subsection{Image Quality Assessment Methods}
\label{sec:related:iqa}

\noindent \textbf{IQA without explicit reasoning} learns an implicit function that directly maps image representations to quality levels without exposing intermediate decision steps.
From early NR-IQA pipelines~\cite{mittal2012no,mittal2012making} that rely on hand-crafted statistics, such as BRISQUE~\cite{mittal2012no}, to modern deep learning methods~\cite{bosse2017deep,talebi2018nima,ke2021musiq}, most approaches still follow the same paradigm: extract a global representation and output a scalar quality score. In this sense, the reasoning process is implicitly manifested by model parameters. With the emergence of VLMs~\cite{liu2023visual,bai2025qwen2}, some attempts have been made to bias the model toward following image quality assessment instructions via Supervised Fine-Tuning (SFT), and then directly output quality ratings~\cite{wu2024qinstruct,wu2023QAlign,zhuadaptive,you2024DQA,you2025teaching,you2024DQAW}. For example, Q-Align~\cite{wu2023QAlign}, Compare2Score~\cite{zhuadaptive}, and DeQA-Score~\cite{you2025teaching} tweak VLMs with sophisticatedly designed quality prompting schemes and achieve significant progress. Such direct scoring is attractive because of its scalability, but limits generalization under distribution shifts. 

\noindent \textbf{IQA with explicit reasoning} encourages the model to decompose complex evaluation tasks into sequential sub-steps. With the emergence of CoT~\cite{wei2022cot} and its variants~\cite{wang2023selfconsistency,zhou2023leasttomost,yao2023tot}, it has become clear that eliciting intermediate steps can impose a positive influence on large models. This motivates a shift from purely score regression to approaches that explicitly verbalize the quality assessment process~\cite{wu2024qinstruct,li2025qinsight,wu2025visualqualityr1}, yielding more explainable IQA outputs. Existing efforts can be classified into three categories based on how reasoning is induced: (i) SFT, such as Q-Instruct~\cite{wu2024qinstruct}, Co-Instruct~\cite{wu2024coinstruct}, and Depict-QA~\cite{you2024DQA}; (ii) RL, e.g., Q-Insight~\cite{li2025qinsight}, VQ-Insight~\cite{zhang2025vqinsight}, VisualQuality-R1~\cite{wu2025visualqualityr1} and Q-Ponder~\cite{cai2025q}; and (iii) active visual tools, including ZoomIQA~\cite{liang2026zoomiqa} and Q-Probe~\cite{cai2025q}. Despite these advances in both reasoning and scoring, the reasoning trajectories predominantly anchor to the textual space. However, representing continuous visual signals via discrete text tokens can entail reductive abstraction~\cite{li2025latent,qin2025covt,ma2025cocova}, which could limit reasoning efficacy in visual tasks.

\subsection{Visual Reasoning in VLMs}
\label{sec:related:lvr}
Visual reasoning refers to performing compositional inference grounded in vision to derive answers beyond direct recognition~\cite{liu2025visual}. Early progress has largely been text-centric, with methods such as CoT prompting~\cite{wei2022cot}, and RL-induced models like Vision-R1~\cite{huang2025visionr1} and Visual-RFT~\cite{liu2025visual} — all of which primarily improve visual question answering by eliciting better textual reasoning traces over fixed visual embeddings. More recent work moves toward interleaved multimodal reasoning by generating visual representations for reasoning steps~\cite{yang2025mirage,qin2025covt} or reintegrating specific visual tokens into reasoning~\cite{chen2025reasoning}. Collectively, these advances aim to enhance the visual awareness of reasoning. A parallel line of work in natural language processing has attempted to explore reasoning in a latent space rather than directly in the token space~\cite{shen2025codi,cheng2024compressed}. COCONUT~\cite{hao2024training} regards hidden states as continuous thoughts to be fed back to the model iteratively, while \cite{goyal2023think} uses pause tokens to allocate extra internal computation without generating visible tokens. More recent approaches have begun to explore latent-space reasoning in VLMs~\cite{li2025latent,li2025latentimplicitvisualreasoning,wang2025monet}. These demonstrate that latent representations can act as an effective reasoning space. Distinctly, \ours is the first work to explore IQA in the latent reasoning space by employing a synergistic two-stage strategy.

\section{Methodology}
\label{sec:methods}

\subsection{Preliminaries}
\label{sec:methods:preliminaries}

\paragraph{\textbf{VLM as token-conditioned decoding.}}
Given an image $x$, a visual encoder $E_v(\cdot)$ produces visual features that are projected into the LLM embedding space via a projector $P(\cdot)$, yielding visual tokens $V=\{v_j\}_{j=1}^{N}=P(E_v(x))$. A conventional VLM then performs autoregressive decoding over a discrete vocabulary conditioned on $V$ and the textual prompt.

\paragraph{\textbf{Latent visual reasoning.}}~\cite{li2025latent} extends decoding with latent segments that alternate with standard text generation. Upon emitting a trigger \texttt{\LVRstart}, the model pauses textual decoding and enters a latent mode, producing a sequence of continuous hidden states
$\mathcal{H}_{\text{latent}}=\{h_t\}_{t=1}^{T}$, where $T$ is the number of hidden-state involved.
After the latent segment, the model emits \texttt{\LVRend} and resumes standard decoding.

\quad\quad\textbf{1) Supervising latent states with visual-token reconstruction.}
To ground latent thoughts, $\mathcal{H}_{\text{latent}}$ is supervised to reconstruct a set of target visual tokens.
Let $\mathcal{I}$ denote indices of visual tokens among ROIs, and let $\phi(\cdot)$ map each $i\!\in\!\mathcal{I}$ to a latent step $\phi(i)\!\in\!\{1,\dots,T\}$. To this end, the MSE loss is employed to optimize the latent reconstruction: 
\begin{equation}
\mathcal{L}_{\text{LVR}}=\frac{1}{|\mathcal{I}|}\sum_{i\in\mathcal{I}}
\left\| g\!\left(h_{\phi(i)}\right)-v_i \right\|_2^2,
\label{eq:llvr}
\end{equation}
where $g(\cdot)$ projects hidden states to the visual-token space. During this process, the model is jointly optimized by the next-token prediction $\mathcal{L}_{\text{NTP}}$ and latent visual reconstruction.  
\begin{equation}
\mathcal{L}_{\text{SFT}}=\mathcal{L}_{\text{NTP}}+\lambda_{\text{lvr}}\mathcal{L}_{\text{LVR}}.
\label{eq:lsft}
\end{equation}
Specifically, the standard cross-entropy loss for next-token prediction is adopted only on the discrete outputs, and the latent reconstruction loss aligns predicted latent states with target visual embeddings.

\quad\quad\textbf{2) Latent GRPO.}
Since the latent segment does not yield token log-probabilities and is treated as an internal continuous computation that conditions generation, GRPO is applied only to discrete output tokens.
Concretely, for each rollout, realized latent states $\hat{\mathcal{H}}_{\text{latent}}$ inside the latent space are recorded and replayed to restore an identical conditioning context. Let $y=\{y_t\}_{t\in\mathcal{T}_{\text{text}}}$ denote the sequence of discrete response tokens, where $\mathcal{T}_{\text{text}}$ is the index set of token-decoding steps (excluding latent steps) and define $T_y = |\mathcal{T}_{\text{text}}|$ as the number of discrete response tokens. The token importance ratio $r_t(\theta)$ under the updated policy $\pi_\theta$ and the rollout policy $\pi_{\text{old}}$ can be written as,
\begin{equation}
r_t(\theta)=
\frac{\pi_\theta\!\left(y_t\,\middle|\,x,\hat{\mathcal{H}}_{\text{latent}},y_{<t}\right)}
{\pi_{\text{old}}\!\left(y_t\,\middle|\,x,\hat{\mathcal{H}}_{\text{latent}},y_{<t}\right)}.
\end{equation}
The latent segment is treated as a fixed conditioning context. For each input, we sample a group of $K$ rollouts and optimize:
\begin{equation}
\mathcal{J}(\theta)=
\mathbb{E}\!\left[
\frac{1}{K}\sum_{k=1}^{K}\sum_{t}
\min\!\Big(r_t(\theta)\hat{A}_t,\,
\text{clip}(r_t(\theta),1-\epsilon,1+\epsilon)\hat{A}_t\Big)
\right]
-\beta \sum_{t}
D_{\text{KL}}\!\left(\pi_\theta\|\pi_{\text{ref}}\right),
\label{eq:J}
\end{equation}
where $\hat{A}_t$ denotes the group-relative advantages, $\pi_{\text{ref}}$ is the SFT reference model, $\epsilon$ is the clipping range, and $\beta$ is the Kullback–Leibler (KL) coefficient.

\begin{figure*}[t]
  \centering
  \includegraphics[width=1\linewidth]{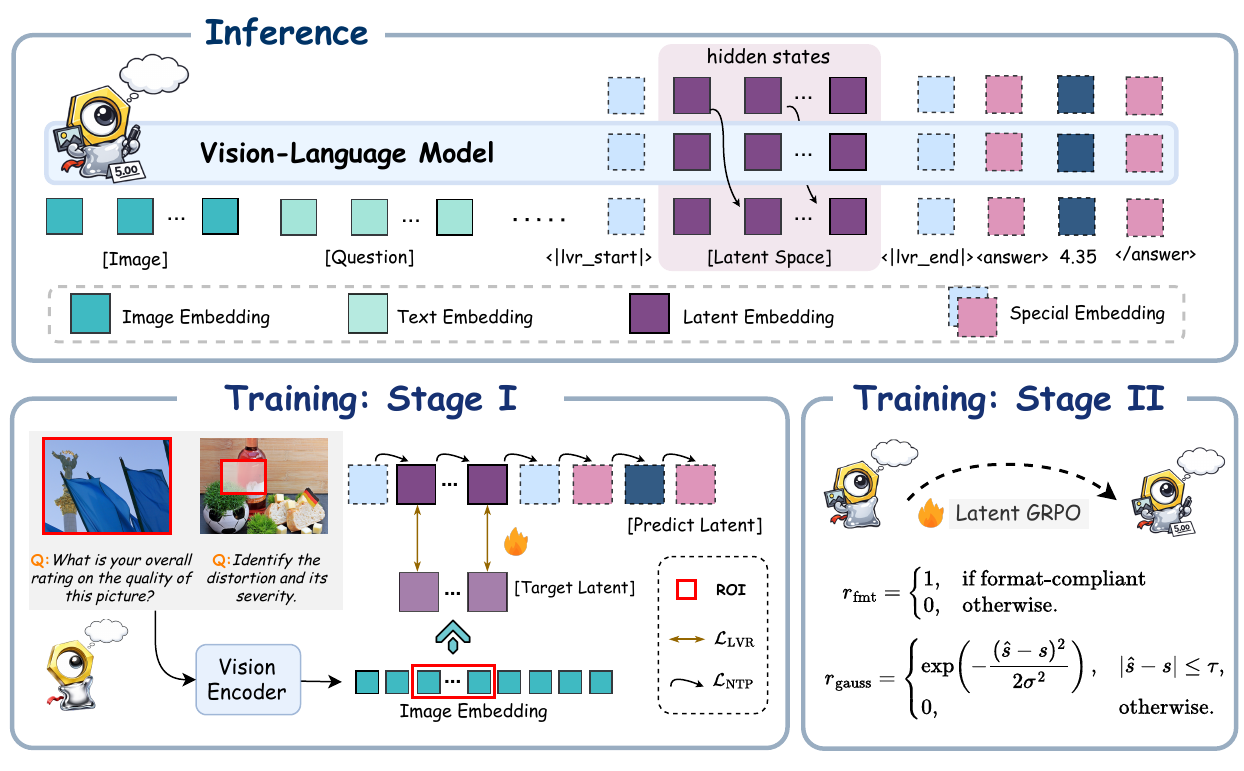}
  \caption{\textbf{Overview of Q-Tacit.} We introduce a compact latent quality-reasoning segment wrapped by special tokens \texttt{<|lvr\_start|>} and \texttt{<|lvr\_end|>}, where the model propagates latent embeddings (i.e., internal hidden states) as a compact quality space instead of generating textual rationales. Then directly outputs the final quality score in \texttt{\answertagl} and \texttt{\answertagr} when a stopping criterion is met.
  }
  
  \label{fig:qtacit_pipeline}
\end{figure*}

\subsection{The Q-Tacit Framework}
\label{sec:methods:qtacit_lvr}

\ours is a reasoning-induced NR-IQA framework that functions in the latent space. As illustrated in~\cref{fig:qtacit_pipeline},
Q-Tacit first learns a latent quality space and latent reasoning patterns via LVR-induced SFT (\cref{sec:stage1}), which are consistently invoked during inference. Next, the reasoning trajectories are calibrated within the latent space via latent GRPO to better assess image quality (\cref{sec:stage2}). We finally elaborate on the data formulation for the two training stages (\cref{sec:method:datacon}). \cref{fig:qtacit_pipeline} shows the latent-reasoning based two-stage paradigm of Q-Tacit, and the details are elaborated below.

\subsubsection{Stage I: Latent Prior Injection via SFT}
\label{sec:stage1}
The main objective of this stage is to form a latent quality space and teach the model basic latent reasoning patterns, including when to invoke latent visual thoughts and how to use them for quality-sensitive evidence reconstruction. As illustrated in Stage I of \cref{fig:qtacit_pipeline}, to build a quality-aware reasoning substrate, we particularly use an ROI-based training strategy, where we enforce the routing of visual evidence relevant to the query through the latent tokens. In practice, we implement this constraint by providing the model with answer-relevant image regions specified by bounding boxes. Concretely, each input query is paired with an answer-relevant image region, which is obtained via cost-free spatial synthesis. 

Given an ROI of the input image, we first convert it into patch-grid coordinates to obtain an index set $\mathcal{I}=\{i_1,\dots,i_{T_v}\}$, where $T_v=|\mathcal{I}|$, and $|\mathcal{I}|$ denotes the length of ROI visual index. The corresponding ROI visual tokens are selected from the full set of image tokens as $V_{\mathrm{ROI}}=\{v_i \mid i\in\mathcal{I}\}$.

\begin{figure*}[!t]
  \centering
  \includegraphics[width=1\linewidth]{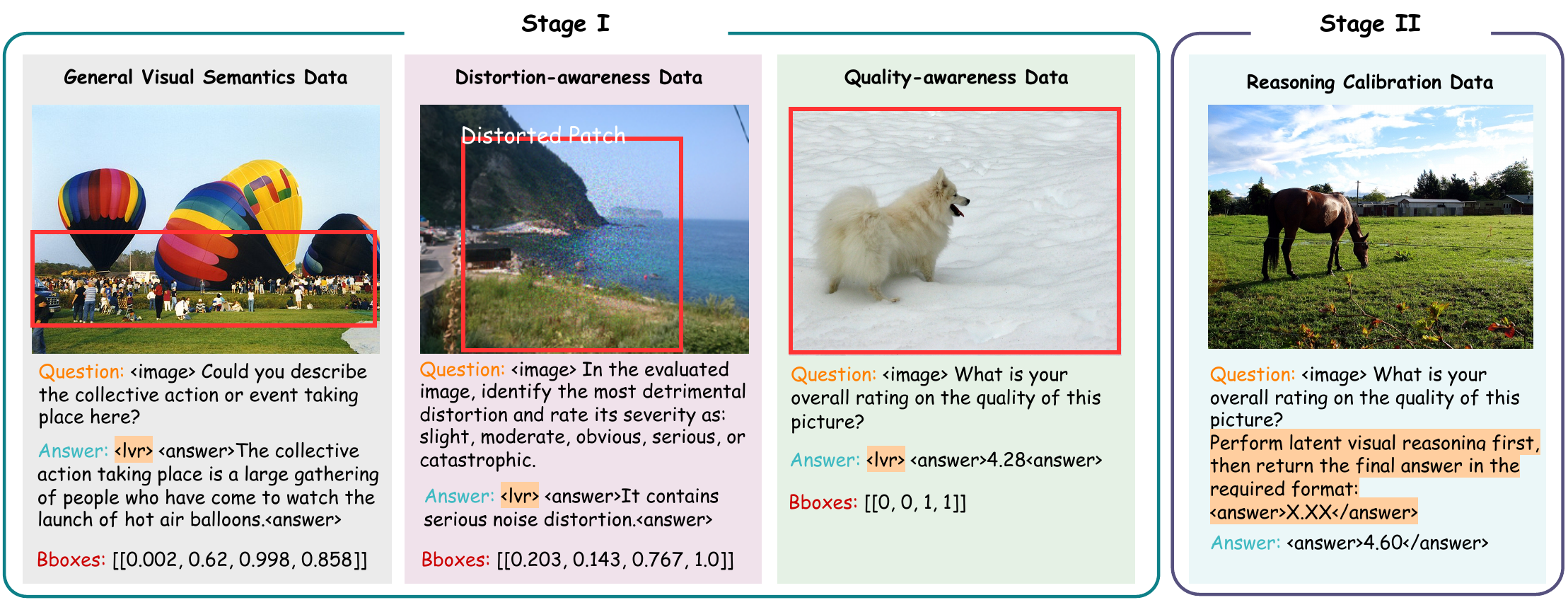}
  \caption{\textbf{Example training data for Q-Tacit.} The left part shows the mixture for constructing the latent space in Stage I. The \texttt{Bboxes} field specifies the ROI to supervise latent reconstruction. \texttt{<lvr>} functions as a placeholder to indicate where the latent segment should be inserted, which will expand to a series of latent span $\texttt{<|lvr\_start|>}\ \texttt{<|lvr|>}... \texttt{<|lvr|>}\texttt{<|lvr\_end|>}$ in practice. The right part is quality-aligned calibration for latent quality reasoning in Stage II.}
  \label{fig:qtacit_data}
\end{figure*}

As shown by the example in \cref{fig:qtacit_data}, each answer begins with the \LVRtag tag, which will expand into a latent segment representing the reasoning process, i.e., \texttt{\LVRstart}\ \texttt{<|lvr|>$_{i}$} \dots \texttt{<|lvr|>$_{T}$} \texttt{\LVRend}, where \texttt{<|lvr|>} functions as a placeholder for the latent slot. The extracted $V_{\text{ROI}}$ is then mapped to the reasoning segment.  In Stage I, we typically set the number of latent slots to match the ROI token count (i.e., $T=T_v$) to provide dense token-level supervision, and align each $i\in\mathcal{I}$ to the corresponding latent step \texttt{<|lvr|>$_{i}$} via the token index projector $\phi(\cdot)$. We then impose a reconstruction objective by treating $V_{\text{ROI}}$ as targets and training the aligned latent slots to reconstruct the ROI visual tokens indicated in \cref{fig:qtacit_pipeline}, optimized by Eq.~\eqref{eq:lsft}. This joint supervision injects quality-relevant ROI evidence into the latent segment and facilitates a quality-aware latent space for subsequent reasoning and calibration.

\subsubsection{Stage II: Quality-aligned Calibration via Latent GRPO}
\label{sec:stage2}
As shown in \cref{fig:qtacit_pipeline}, while Stage I is trained with supervised objectives under teacher-forcing, Stage II calibrates the end-to-end scoring behavior via Reinforcement Learning (RL) on the reasoning calibration data (Fig.~\ref{fig:qtacit_data}, right), aligning the model with human-rated quality scores.
In Stage II, we use a fixed latent-step budget $T$ because ROI token counts are unavailable for real IQA data. This fixed budget stabilizes policy optimization by keeping trajectory length and compute consistent, and making KL regularization easier to control.

\paragraph{\textbf{Reward design.}}
Since the final output is evaluated solely by its scalar score, rewards are computed from (i) format validity and (ii) the parsed score $\hat{s}$ between \answertagl\ and \answertagr. 

\quad\quad\textbf{1) Format compliance reward $r_{\text{fmt}}$.}
We use a binary reward to enforce structural validity:
$r_{\text{fmt}}=1$ if the response contains a valid latent segment bounded by
\texttt{\LVRstart}$\cdots$\texttt{\LVRend} and a parsable \texttt{\answertagl}\texttt{\answertagr} with a numeric \texttt{value} inside; otherwise, $r_{\text{fmt}}=0$.
This term prevents the policy from drifting away from the intended hybrid interface during RL.

\quad\quad\textbf{2) Dense quality scoring reward $r_{\text{gauss}}$.}
To avoid sparse supervision, we shape score accuracy with a continuous Gaussian kernel:
\begin{equation}
r_{\text{gauss}}=
\begin{cases}
\exp\!\left(-\frac{(\hat{s}-s)^2}{2\sigma^2}\right), & \text{if } |\hat{s}-s|\le \tau\\
0, & \text{otherwise},
\end{cases}
\label{eq:rgauss}
\end{equation}
where $s$ is the ground-truth quality score, $\sigma$ controls tolerance, and $\tau$ truncates overly inaccurate predictions to reduce noisy gradients early in training. This yields smooth, bounded feedback (in $[0,1]$).

\paragraph{\textbf{Total reward and advantage.}}
We combine these two terms as
$r_{\text{total}}=r_{\text{gauss}}+r_{\text{fmt}}$, where $r_{\text{gauss}}\in[0,1]$ and $r_{\text{fmt}}\in\{0,1\}$.
Following GRPO, for each input, we sample $K$ rollouts, compute group-relative advantages by centering rewards within the group, and optimize the objective in~\cref{eq:J}.

\subsubsection{\textbf{Inference and output format.}}
At inference, we implement a two-mode decoding procedure. Once \texttt{<|lvr\_start|>} is produced, the model enters the latent reasoning mode and iteratively computes a sequence of latent hidden states $\{h_t\}_{t=1}^{T}$, and then exits the latent segment by generating \texttt{<|lvr\_end|>}. We enforce a fixed latent-step budget $T$. If \texttt{<|lvr\_end|>} is not generated before reaching $T$, we append it to close the segment to ensure a consistent reasoning length. The final visible output is a scalar score enclosed by \texttt{\answertagl} and \texttt{\answertagr}, as shown in~\cref{fig:qtacit_pipeline}.

\subsection{Data Construction}
\label{sec:method:datacon}
Data composition diversifies for the two stages, which is detailed in \cref{fig:qtacit_data}. In \textbf{Stage~I}, we first incorporate queries on high-level visual recognition to build a general visual semantic substrate for the latent reasoning space, and further include IQA-specific queries to instill a quality-centric prior. Specifically, for distortion awareness, each input consists of a prompt and an image containing a distorted patch via a randomly generated ROI mask. There are five distortion types, including ``noise'', ``compression'', ``blur'', ``photometric'', and ``null'' (no distortion). Each distortion has five severity levels, including ``slight'', ``moderate'', ``obvious'', ``serious'', and ``catastrophic''. We mix diversified prompts to reduce rigid model responses. For quality-awareness, human-annotated quality scores encourage quality rating prior into the reasoning space. As such, these multi-purpose data sources are complementary, where the high-level vision part stabilizes generic latent thinking behaviors, whereas the IQA-specific queries anchor the latent space to the target perceptual quality dimension. In \textbf{Stage II}, we employ queries on image quality ratings to calibrate the model's reasoning trajectories towards human perception via latent GRPO. More detailed data usage is elaborated in~\cref{sec:implement}.

\section{Experiments}
\label{sec:exp}
\subsection{Implementation Details}
\label{sec:implement}
We adopt Qwen2.5-VL-7B-Instruct~\cite{bai2025qwen2} as the backbone model. The visual encoder and multimodal projector are kept frozen, with only LLM parameters updated throughout training. All training runs on 16$\times$NVIDIA GH200 120GB GPUs~\cite{mcintosh2024isambard} with a total batch-size of 128.

For \textbf{Stage~I}, we randomly sampled 13K data from the Visual CoT dataset~\cite{shao2024visual} for general visual semantics. We also sampled 40K images from KADIS-700K~\cite{deepfl-iqa} for distortion awareness, and employed the training split of KonIQ-10K~\cite{hosu2020koniq} for quality awareness, containing 7K images following \cite{you2025teaching}. The adaptive multimodal data-packing strategy~\cite{chen2024expanding} was employed to support dynamic batching. The learning rate is $1\times10^{-5}$, and $\lambda_{\text{lvr}}$ is set to 0.1. It takes about 2 hours to complete 4 epochs for convergence. \textbf{Stage~II} performs quality-aligned quality calibration on the training split of KonIQ-10K. The group size for rollouts is set to 8. The weight of the KL divergence penalty $\beta$ is $1\times10^{-3}$, the budget $T$ is set to 8, $\sigma = 0.5$, and $\tau=1$. Besides, the AdamW optimizer~\cite{loshchilov2019decoupled} is employed with an initial learning rate of $1\times10^{-6}$. The model is trained for 10 epochs for this phase, taking approximately 25 hours.

\subsection{Competing Models and Evaluation Datasets}

We report IQA performance across four types of datasets:
(1) in-the-wild: KonIQ-10K~\cite{hosu2020koniq}, SPAQ~\cite{fang2020perceptual}, and LIVE-Wild~\cite{ghadiyaram2015live};
(2) synthetic distortions: KADID~\cite{lin2019kadid} and CSIQ~\cite{larson2010most};
(3) model-processed distortions: PIPAL~\cite{jinjin2020pipal}; and
(4) AI-generated: AGIQA~\cite{li2023agiqa}. MOSs of these datasets are normalized to the range of [1,5], with the variances normalized accordingly. We compare \ours with handcrafted
methods NIQE~\cite{mittal2012making} and BRISQUE~\cite{mittal2012no}; deep-learning methods including NIMA~\cite{talebi2018nima}, 
MUSIQ~\cite{ke2021musiq}, CLIP-IQA+~\cite{wang2023exploring}, and ManIQA~\cite{yang2022maniqa}; and recent VLM-based methods, including Compare2Score~\cite{zhuadaptive}, Q-Align~\cite{wu2023QAlign}, DeQA-Score~\cite{you2025teaching}, Q-Insight~\cite{li2025qinsight}, VisualQuality-R1~\cite{wu2025visualqualityr1} and supervised fine-tuned Qwen-2.5-VL-7B~\cite{bai2025qwen2}. These VLM-based methods contain approximately 7B parameters. For a fair comparison, all compared models are trained on KonIQ-10K~\cite{hosu2020koniq}, unless explicitly stated otherwise. We adopt PLCC and SRCC to measure performance following~\cite{yang2022maniqa,wu2023QAlign}. 

\begin{table*}[t]
\renewcommand{\arraystretch}{1}
\fontsize{8}{8}\selectfont
\setlength{\tabcolsep}{3pt}
\centering
\caption{\textbf{PLCC / SRCC results of NR-IQA models trained on KonIQ-10K}. Exceptions marked with $^\dagger$ use a multi-task training set in \cite{li2025qinsight} for GRPO calibration. Top two results are highlighted in \best{bold} and \secondbest{underlined}, respectively.}
\label{tab:score_both}

\begingroup
\providecommand{\color}[2][]{ } 

\newlength{\MetW}
\setlength{\MetW}{-3pt}
\addtolength{\MetW}{2\tabcolsep} 

\newlength{\PairHt}
\setlength{\PairHt}{5.70ex} 

\newcommand{\PairBG}[1]{%
  \noalign{%
    \begingroup
    \rlap{%
      \kern\MetW
      \color{#1}\rule{330pt}{\PairHt}
    }%
    \vskip-\PairHt
    \endgroup
  }%
}

\resizebox{\linewidth}{!}{
\begin{tabular}{l cccccccc}
\toprule
\textbf{Methods} & \textbf{KonIQ-10K} & \textbf{SPAQ} & \textbf{KADID} & \textbf{PIPAL} & \textbf{LIVEW} & \textbf{AGIQA} & \textbf{CSIQ} & \textbf{AVG.} \\
\midrule

\multicolumn{9}{l}{\textbf{Handcrafted}}\\[-0.25ex]
\PairBG{gray!8}
NIQE~\cite{mittal2012making} & 0.533 & 0.679 & 0.468 & 0.195 & 0.493 & 0.560 & 0.718 & 0.521 \\
(SPL 2012) & \srcc{0.530} & \srcc{0.664} & \srcc{0.405} & \srcc{0.161} & \srcc{0.449} & \srcc{0.533} & \srcc{0.628} & \srcc{0.481} \\
\PairBG{gray!2}
BRISQUE~\cite{mittal2012no} & 0.225 & 0.490 & 0.429 & 0.267 & 0.361 & 0.541 & 0.740 & 0.436 \\
(TIP 2012) & \srcc{0.226} & \srcc{0.406} & \srcc{0.356} & \srcc{0.232} & \srcc{0.313} & \srcc{0.497} & \srcc{0.556} & \srcc{0.369} \\
\midrule

\multicolumn{9}{l}{\textbf{Non-VLM Deep-learning}}\\[-0.25ex]
\PairBG{gray!8}
NIMA~\cite{talebi2018nima} & 0.896 & 0.838 & 0.532 & 0.390 & 0.814 & 0.715 & 0.695 & 0.697 \\
(TIP 2018) & \srcc{0.859} & \srcc{0.856} & \srcc{0.535} & \srcc{0.399} & \srcc{0.771} & \srcc{0.654} & \srcc{0.649} & \srcc{0.675} \\
\PairBG{gray!2}
MUSIQ~\cite{ke2021musiq} & 0.924 & 0.868 & 0.575 & 0.431 & 0.789 & 0.722 & 0.771 & 0.726 \\
(ICCV 2021) & \srcc{0.929} & \srcc{0.863} & \srcc{0.556} & \srcc{0.431} & \srcc{0.830} & \srcc{0.630} & \srcc{0.710} & \srcc{0.707} \\
\PairBG{gray!8}
CLIP-IQA+~\cite{wang2023exploring} & 0.909 & 0.866 & 0.653 & 0.427 & 0.832 & 0.736 & 0.772 & 0.742 \\
(AAAI 2023) & \srcc{0.895} & \srcc{0.864} & \srcc{0.654} & \srcc{0.419} & \srcc{0.805} & \srcc{0.685} & \srcc{0.719} & \srcc{0.720} \\
\PairBG{gray!2}
ManIQA~\cite{yang2022maniqa} & 0.849 & 0.768 & 0.499 & 0.457 & 0.849 & 0.723 & 0.623 & 0.681 \\
(CVPR 2022) & \srcc{0.834} & \srcc{0.758} & \srcc{0.465} & \srcc{0.452} & \srcc{0.832} & \srcc{0.636} & \srcc{0.627} & \srcc{0.658} \\
\midrule

\multicolumn{9}{l}{\textbf{VLM-based}}\\[-0.25ex]
\PairBG{gray!8}
Compare2Score~\cite{zhuadaptive} & 0.923 & 0.867 & 0.500 & 0.354 & 0.786 & 0.777 & 0.735 & 0.706 \\
(NeurIPS 2024) & \srcc{0.910} & \srcc{0.860} & \srcc{0.453} & \srcc{0.342} & \srcc{0.772} & \srcc{0.671} & \srcc{0.705} & \srcc{0.673} \\
\PairBG{gray!2}
Qwen-2.5-VL-7B~\cite{bai2025qwen2} & 0.889 & 0.874 & 0.668 & 0.473 & 0.734 & 0.813 & 0.674 & 0.732 \\
(Arxiv 2025) & \srcc{0.866} & \srcc{0.875} & \srcc{0.663} & \srcc{0.442} & \srcc{0.728} & \srcc{0.739} & \srcc{0.650} & \srcc{0.709} \\
\PairBG{gray!8}
Q-Align~\cite{wu2023QAlign} & \secondbest{0.941} & 0.886 & 0.674 & 0.403 & 0.853 & 0.772 & 0.671 & 0.705 \\
(ICML 2024) & \srcc{\secondbest{0.940}} & \srcc{0.887} & \srcc{0.684} & \srcc{0.419} & \srcc{0.860} & \srcc{0.735} & \srcc{0.737} & \srcc{0.752} \\
\PairBG{gray!2}
DeQA-Score~\cite{you2025teaching} & \best{0.953} & 0.895 & 0.694 & 0.472 & \secondbest{0.892} & 0.809 & 0.787 & 0.786 \\
(CVPR 2025) & \srcc{\best{0.941}} & \srcc{0.896} & \srcc{0.687} & \srcc{0.478} & \srcc{\best{0.879}} & \srcc{0.729} & \srcc{0.744} & \srcc{0.765} \\
\PairBG{gray!8}
VisualQuality-R1~\cite{wu2025visualqualityr1} & 0.924 & 0.894 & 0.701 & 0.469 & 0.873 & 0.822 & 0.743 & 0.775 \\
(NeurIPS 2025) & \srcc{0.915} & \srcc{0.895} & \srcc{0.694} & \srcc{0.446} & \srcc{0.841} & \srcc{\best{0.773}} & \srcc{0.689} & \srcc{0.750} \\
\PairBG{gray!2}
Q-Insight~\cite{li2025qinsight} & 0.918 & 0.903 & 0.702 & 0.458 & 0.870 & 0.816 & 0.685 & 0.765 \\
(NeurIPS 2025) & \srcc{0.895} & \srcc{0.899} & \srcc{0.702} & \srcc{0.435} & \srcc{0.839} & \srcc{0.766} & \srcc{0.640} & \srcc{0.739} \\
\PairBG{gray!8}
Q-Insight$^\dagger$~\cite{li2025qinsight} & 0.933 & 0.907 & 0.742 & 0.486 & \best{0.893} & 0.811 & 0.870 & 0.806 \\
(NeurIPS 2025) & \srcc{0.916} & \srcc{0.905} & \srcc{0.736} & \srcc{0.474} & \srcc{0.865} & \srcc{0.764} & \srcc{0.824} & \srcc{0.783} \\
\PairBG{gray!2}
\ours & 0.923 & \secondbest{0.911} & \secondbest{0.754} & \secondbest{0.488} & 0.875 & \secondbest{0.823} & \secondbest{0.879} & \secondbest{0.808} \\
\textbf{(Ours)} & \srcc{0.914} & \srcc{\secondbest{0.923}} & \srcc{\secondbest{0.749}} & \srcc{\best{0.493}} & \srcc{0.832} & \srcc{0.765} & \srcc{\secondbest{0.842}} & \srcc{\secondbest{0.788}} \\
\PairBG{gray!8}
\ours$^\dagger$ & 0.925 & \best{0.913} & \best{0.766} & \best{0.511} & 0.890 & \best{0.828} & \best{0.885} & \best{0.817} \\
\textbf{(Ours)} & \srcc{0.914} & \srcc{\best{0.924}} & \srcc{\best{0.760}} & \srcc{\secondbest{0.489}} & \srcc{\secondbest{0.866}} & \srcc{\secondbest{0.769}} & \srcc{\best{0.850}} & \srcc{\best{0.796}} \\
\bottomrule
\end{tabular}}
\endgroup
\end{table*}

\subsection{Main Results}

We compare the quality prediction performance of \ours with other IQA methods in~\cref{tab:score_both} with the following key observations. \underline{Firstly}, VLM-based methods outperform traditional and discriminating deep-learning-based methods, with Qwen-2.5-VL exhibiting strong zero-shot quality scoring ability, indicating the competence of VLMs. \underline{Secondly}, reasoning-induced methods such as Q-Insight, VisualQuality-R1, and Q-Tacit, offer better cross-dataset generalization compared to pure SFT-based methods such as Q-Align and DeQA-Score. \underline{Thirdly}, Q-Tacit achieves the best performance on average, indicating that reasoning in latent space aligns better with human perception of quality than text-centric quality reasoning. Moreover, Q-Tacit enables scalable calibration of reasoning trajectories via Stage II's training. We exploit this and train Q-Tacit$^\dagger$ on the training data of Q-Insight$^\dagger$~\cite{li2025qinsight}, which contains additional image distortion perception data. As shown in the last two lines of \cref{tab:score_both}, Q-Tacit$^\dagger$ yields consistent gains across all seven datasets and achieves state-of-the-art performance among all compared models. This reflects the effectiveness of latent-space reasoning supervision for learning transferable quality representations. Overall, these provide empirical support for the precondition that a compact latent space can operate more directly on perceptual evidence, thereby improving generalization ability.

\begin{figure*}[!t]
  \centering
  \includegraphics[width=1\linewidth]{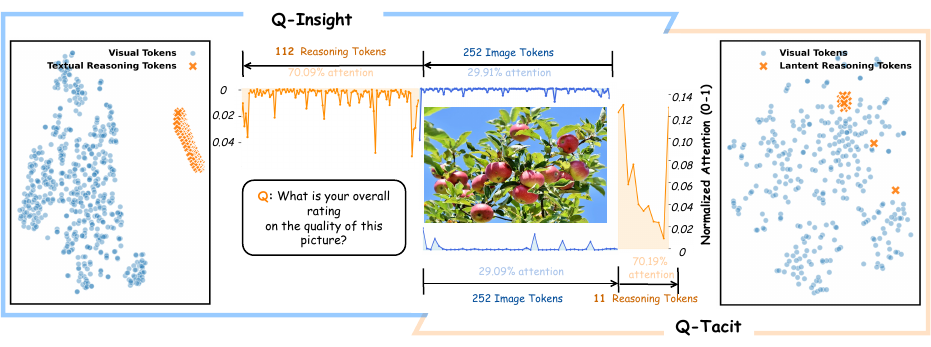}
  \caption{\textbf{Token-space coupling and attention weight distribution during quality scoring: Q-Insight vs. Q-Tacit.}
  Left/Right: t-SNE visualization of visual and reasoning tokens for Q-Insight/Q-Tacit. Middle: normalized attention weights over the image and reasoning tokens after the quality score is generated. Q-Tacit obtained similar attention allocated to reasoning with only 11 tokens (vs. 112 for Q-Insight). }

\label{fig:attn_score}
\end{figure*}

To further investigate the mechanism underlying Q-Tacit's effectiveness, we compare the attention patterns of Q-Insight and Q-Tacit during scoring. Specifically, we normalize the \underline{image} and \underline{reasoning} tokens after the quality score is generated and visualize them. As illustrated in \cref{fig:attn_score}, quality score generation predominantly attends to reasoning tokens for both methods. Q-Tacit assigns a larger fraction of attention to its reasoning tokens than the text-centric Q-Insight, with far fewer reasoning tokens ($\approx$10\%). Besides, t-SNE~\cite{van2008visualizing} visualizations show that Q-Tacit’s reasoning tokens are more interleaved with visual tokens, suggesting tighter coupling to visual embeddings. This indicates that the latent space compactly encodes quality-relevant visual evidence, further supporting latent continuous space as a favorable alternative to purely language-centric reasoning for IQA.

\begin{figure*}[t]
  \centering
  \includegraphics[width=\linewidth]{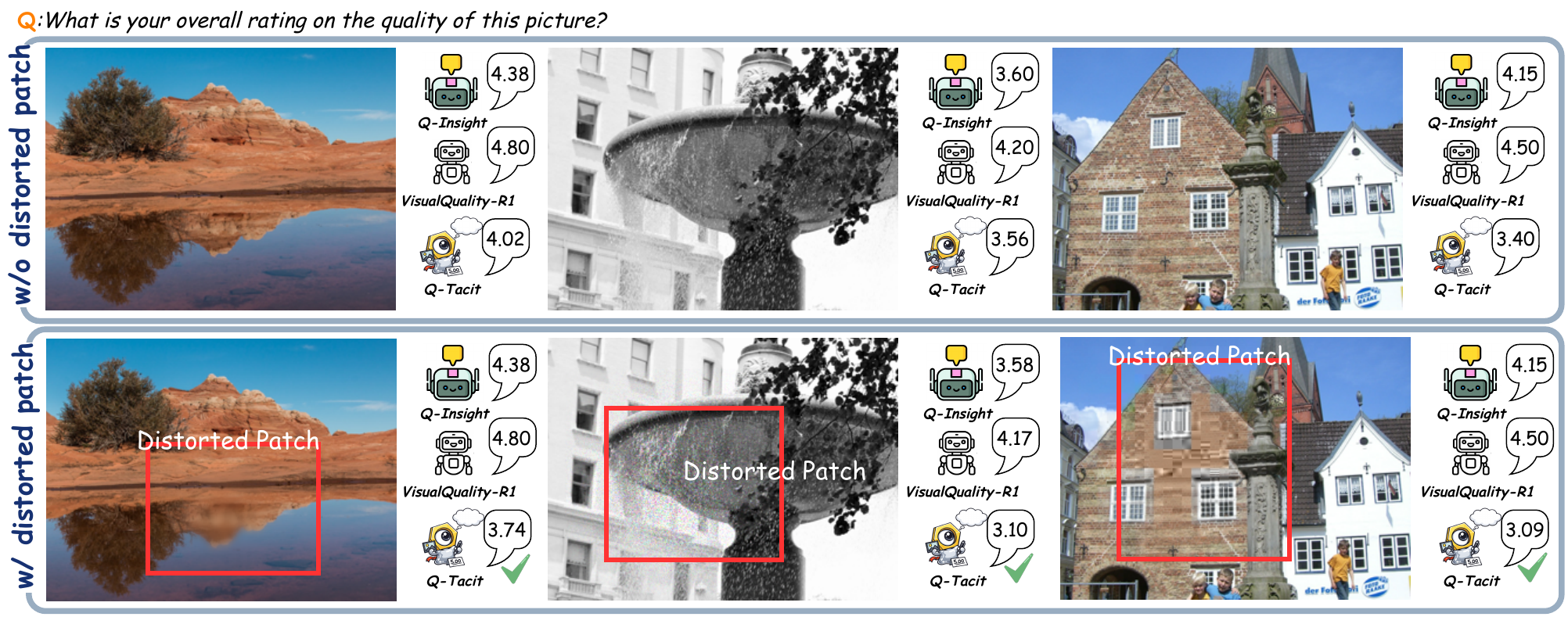}
  \caption{\textbf{Sensitivity to localized distortions.}
  We compare Q-Insight, VisualQuality-R1, and Q-Tacit by reporting predicted quality scores for each image (w/o distorted patch, top) and its counterpart with a localized corruption (w/ distorted patch, bottom). Blur, noise, and compression are injected within the red box (left to right). Q-Tacit consistently lowers its predicted scores in response to the local distortion.}
  \label{fig:patchcase}
\end{figure*}

Moreover, we probe whether the proposed Q-Tacit can reason with concrete visual quality evidence and respond to subtle local degradations. As shown in \cref{fig:patchcase}, we manually insert a distorted patch into otherwise unchanged images, and compare the predicted quality of Q-Tacit with other two reasoning-induced methods, Q-Insight~\cite{li2025qinsight} and VisualQuality-R1~\cite{wu2025visualqualityr1} using their original reasoning prompt templates. Q-Tacit consistently lowers its predicted scores in response to the local distortion, whereas Q-Insight and VisualQuality-R1 are markedly less sensitive. This indicates that Q-Tacit’s latent reasoning is more visual-aware than text-centric reasoning.


\subsection{Ablation Studies}

\paragraph{\textbf{Study1: Effect of latent reasoning.}} We first verify that the gain of Q-Tacit stems from the proposed latent-space reasoning mode, rather than from additional training beyond RL or from RL alone.
We compare (i) Stage I-only, and (ii) Stage I+Stage II without the LVR mode. As shown in \cref{tab:S1}, the non-LVR variant matches in-domain performance on KonIQ-10K, but exhibits noticeably weaker OOD generalization. In contrast, enabling latent reasoning consistently improves cross-dataset robustness. These clearly indicate that the observed gains are primarily driven by latent-space reasoning, whereas neither additional training nor RL alone can account for the full improvements.

\begin{table*}[t]
\renewcommand{\arraystretch}{1.1}
\setlength{\tabcolsep}{3pt}
\centering
\caption{\textbf{Ablation on the effect of latent reasoning pipeline (PLCC/SRCC).}
We compare Q-Tacit with two variants: \emph{Stage I-only} and \emph{Stage I+II w/o LVR}. Top two results are highlighted in \best{bold} and \secondbest{underlined}, respectively.}
\label{tab:S1}

\resizebox{\linewidth}{!}{%
\begin{NiceTabular}{c|ccccccc}
\CodeBefore
  \tikz \fill [gray!10] (1-|1) rectangle (2-|9);
\Body
\hline
\textbf{Training Strategy} & \textbf{KonIQ-10K} & \textbf{SPAQ} & \textbf{KADID} & \textbf{PIPAL} & \textbf{LIVEW} & \textbf{AGIQA} & \textbf{CSIQ} \\
\hline
Stage I only
& 0.901\srcc{0.899}
& \secondbest{0.904}\srcc{\secondbest{0.901}}
& 0.733\srcc{0.731}
& 0.473\srcc{0.467}
& 0.853\srcc{0.811}
& \secondbest{0.812}\srcc{0.737}
& 0.839\srcc{0.802} \\

Stage I+II w/o LVR
& \secondbest{0.920}\srcc{\secondbest{0.912}}
& 0.901\srcc{0.898}
& \secondbest{0.736}\srcc{\secondbest{0.735}}
& \secondbest{0.476}\srcc{\secondbest{0.475}}
& \secondbest{0.871}\srcc{\best{0.834}}
& 0.809\srcc{\secondbest{0.759}}
& \secondbest{0.867}\srcc{\secondbest{0.825}} \\

Q-Tacit (Ours)
& \best{0.923}\srcc{\best{0.914}}
& \best{0.911}\srcc{\best{0.923}}
& \best{0.754}\srcc{\best{0.749}}
& \best{0.488}\srcc{\best{0.493}}
& \best{0.875}\srcc{\secondbest{0.832}}
& \best{0.823}\srcc{\best{0.765}}
& \best{0.879}\srcc{\best{0.842}} \\
\hline
\end{NiceTabular}%
}
\end{table*}

\begin{table*}[!t]
\renewcommand{\arraystretch}{1.1}
\setlength{\tabcolsep}{3pt}
\centering
\caption{\textbf{Ablation on the latent space composition (PLCC/SRCC).}
We vary the Stage I training data on three subsets ($\checkmark$ indicates included data), while keeping Stage II identical across rows. Top two results are highlighted in \best{bold} and \secondbest{underlined}, respectively.}
\label{tab:S2}

\resizebox{\linewidth}{!}{%
\begin{NiceTabular}{ccc|ccccccc}
\CodeBefore
  \tikz \fill [gray!10] (1-|1) rectangle (2-|11);
\Body
\hline
\textbf{\makecell[c]{General\\Vision}} & \textbf{\makecell[c]{Quality\\awareness}} & \textbf{\makecell[c]{Distortion\\awareness}}
&\textbf{KonIQ-10K} &\textbf{SPAQ} &\textbf{KADID} &\textbf{PIPAL} &\textbf{LIVEW} &\textbf{AGIQA} &\textbf{CSIQ} \\
\hline
$\checkmark$ & & &
0.901\srcc{0.889} &
0.895\srcc{0.892} &
0.701\srcc{0.699} &
0.453\srcc{0.437} &
0.869\srcc{\secondbest{0.832}} &
0.820\srcc{\best{0.769}} &
0.688\srcc{0.655} \\

$\checkmark$ & & $\checkmark$  &
\secondbest{0.920}\srcc{\best{0.915}} &
\secondbest{0.909}\srcc{\secondbest{0.918}} &
\secondbest{0.753}\srcc{\best{0.749}} &
\secondbest{0.482}\srcc{\secondbest{0.491}} &
\best{0.875}\srcc{\best{0.835}} &
\best{0.825}\srcc{0.761} &
\secondbest{0.875}\srcc{\secondbest{0.838}} \\

$\checkmark$ & $\checkmark$ &  &
0.915\srcc{0.899} &
0.903\srcc{0.901} &
0.710\srcc{\secondbest{0.709}} &
0.461\srcc{0.450} &
\secondbest{0.870}\srcc{\best{0.835}} &
0.818\srcc{\secondbest{0.765}} &
0.732\srcc{0.679} \\

$\checkmark$ & $\checkmark$ & $\checkmark$ &
\best{0.923}\srcc{\secondbest{0.914}} &
\best{0.911}\srcc{\best{0.923}} &
\best{0.754}\srcc{\best{0.749}} &
\best{0.488}\srcc{\best{0.493}} &
\best{0.875}\srcc{\secondbest{0.832}} &
\secondbest{0.823}\srcc{\secondbest{0.765}} &
\best{0.879}\srcc{\best{0.842}} \\
\hline
\end{NiceTabular}
}
\end{table*}

\paragraph{\textbf{Study2: Effect of latent space composition.}} 
\cref{tab:S2} shows how Stage I forms a usable latent visual reasoning space under three data mixtures: (i) general vision, (ii) general vision + distortion awareness, and (iii) general vision + quality awareness. 
It can be observed that adding distortion awareness is the most critical to Q-Tacit, since it grounds the general visual latent space to specific quality evidence. Meanwhile, excluding quality awareness introduces a subtle performance drop, but it can slow the subsequent Stage II training in practice. Overall, the full mixture performs best, suggesting a complementary role of the components, where high-level data stabilizes generic latent reasoning, while IQA-specific perception and scoring anchor the latent space to perceptual quality, thus improving generalization. 

\paragraph{\textbf{Study3: Effect of reward function in latent GRPO.}} To evaluate the contribution of the proposed reward function described in Sec.~\ref{sec:stage2}, we compare it with two alternatives: Q-Insight~\cite{li2025qinsight} and R2RL~\cite{wu2025visualqualityr1}. We also assess the sensitivity to the threshold $\tau$ in $r_{gauss}$ (\cref{eq:rgauss}). As reported in \cref{tab:S2}, our proposed reward function yields higher performance compared to the other two, suggesting that our reward can better guide reasoning calibration. Besides, Q-Tacit achieves relatively stable performance across different threshold values $\tau$, but it should not be too loose or too tight.

\begin{table*}[t] \renewcommand{\arraystretch}{1.1} 
\setlength{\tabcolsep}{3pt} \centering \caption{\textbf{Ablation on reward design (PLCC/SRCC).} Q-Insight and R2RL are two different reward design, and $\tau$ is the threshold in our dense quality scoring reward. Top two results are highlighted in \best{bold} and \secondbest{underlined}, respectively.} 
\label{tab:S3}

\resizebox{\linewidth}{!}{%
\begin{NiceTabular}{l| ccccccc}
\CodeBefore
  \tikz \fill [gray!10] (1-|1) rectangle (2-|9);
\Body
\hline
\textbf{Reward} & \textbf{KonIQ-10K} & \textbf{SPAQ} & \textbf{KADID} & \textbf{PIPAL} & \textbf{LIVEW} & \textbf{AGIQA} & \textbf{CSIQ} \\
\hline
Q-Insight~\cite{li2025qinsight}
& \best{0.928}\srcc{\secondbest{0.908}}
& 0.901\srcc{0.903}
& 0.743\srcc{0.735}
& 0.481\srcc{0.479}
& 0.855\srcc{0.812}
& 0.812\srcc{0.758}
& 0.856\srcc{\secondbest{0.834}} \\

R2RL~\cite{wu2025visualqualityr1}
& 0.922\srcc{0.907}
& 0.905\srcc{\secondbest{0.913}}
& \best{0.756}\srcc{0.745}
& \secondbest{0.485}\srcc{0.481}
& \secondbest{0.859}\srcc{\secondbest{0.827}}
& 0.815\srcc{\secondbest{0.761}}
& 0.863\srcc{0.827} \\

\hline
$\tau=0.35$ 
& \secondbest{0.924}\srcc{\best{0.914}}
& \secondbest{0.908}\srcc{\secondbest{0.913}}
& 0.752\srcc{\best{0.749}}
& 0.479\srcc{0.475}
& \secondbest{0.859}\srcc{0.815}
& \secondbest{0.820}\srcc{0.760}
& 0.864\srcc{0.823} \\

$\tau=2$ 
& 0.921\srcc{\secondbest{0.908}}
& 0.901\srcc{0.905}
& 0.753\srcc{\secondbest{0.747}}
& 0.483\srcc{\secondbest{0.487}}
& 0.855\srcc{0.819}
& 0.814\srcc{0.750}
& \secondbest{0.870}\srcc{0.833} \\

$\tau=1$ (Ours)
& 0.923\srcc{\best{0.914}}
& \best{0.911}\srcc{\best{0.923}}
& \secondbest{0.754}\srcc{\best{0.749}}
& \best{0.488}\srcc{\best{0.493}}
& \best{0.875}\srcc{\best{0.832}}
& \best{0.823}\srcc{\best{0.765}}
& \best{0.879}\srcc{\best{0.842}} \\
\hline
\end{NiceTabular}
}
\end{table*}

\paragraph{\textbf{Study4: Effect of latent reasoning budget $T$.}} We vary the latent reasoning budget $T$ aside from the default $T=8$, which controls the number of latent update steps. \cref{tab:S4} shows that either reducing $T$ to 4 or increasing it to 16 has only a marginal effect, offering an advantageous trade-off between computational complexity and performance.

\begin{table*}[t]
\renewcommand{\arraystretch}{1.1}
\setlength{\tabcolsep}{3pt}
\centering
\caption{\textbf{Ablation on latent reasoning budget $T$ (PLCC/SRCC).}
We vary the latent-segment length $T\in\{4,16\}$ and compare with the default setting $T=8$. Top two results are highlighted in \best{bold} and \secondbest{underlined}, respectively.}
\label{tab:S4}

\resizebox{\linewidth}{!}{%
\begin{NiceTabular}{l| ccccccc}
\CodeBefore
  \tikz \fill [gray!10] (1-|1) rectangle (2-|9);
\Body
\hline
\textbf{Budget} & \textbf{KonIQ-10K} & \textbf{SPAQ} & \textbf{KADID} & \textbf{PIPAL} & \textbf{LIVEW} & \textbf{AGIQA} & \textbf{CSIQ} \\
\hline
$T=4$
& \secondbest{0.920}\srcc{\secondbest{0.913}}
& 0.910\srcc{\best{0.924}}
& 0.745\srcc{0.740}
& 0.481\srcc{0.489}
& \secondbest{0.871}\srcc{\secondbest{0.829}}
& 0.812\srcc{0.749}
& \secondbest{0.864}\srcc{0.826} \\

$T=16$
& 0.919\srcc{0.910}
& \best{0.912}\srcc{0.922}
& \best{0.755}\srcc{\best{0.751}}
& \secondbest{0.487}\srcc{\secondbest{0.492}}
& 0.870\srcc{0.828}
& \best{0.824}\srcc{\best{0.766}}
& \best{0.879}\srcc{\best{0.843}} \\

$T=8$ (Ours)
& \best{0.923}\srcc{\best{0.914}}
& \secondbest{0.911}\srcc{\secondbest{0.923}}
& \secondbest{0.754}\srcc{\secondbest{0.749}}
& \best{0.488}\srcc{\best{0.493}}
& \best{0.875}\srcc{\best{0.832}}
& \secondbest{0.823}\srcc{\secondbest{0.765}}
& \best{0.879}\srcc{\secondbest{0.842}} \\
\hline
\end{NiceTabular}
}
\end{table*}

\section{Conclusions}
In this paper, we introduce \ours, a reasoning-induced NR-IQA framework that enhances the reasoning processes for image quality assessment in a continuous latent visual space.
Unlike prior VLM-based IQA approaches that mainly leverage long textual rationales, \ours frames quality assessment as latent evidence reconstruction and aggregation, enabling compact, visually grounded reasoning with a strict score-only output.
Our two-stage recipe injects structural visual quality priors into the latent space via multi-task SFT and further calibrates latent reasoning trajectories with latent GRPO, improving robustness under diverse distortion regimes.
Extensive experiments across authentic, synthetic, model-processed, and AI-generated benchmarks demonstrate strong cross-dataset generalization and competitive in-domain performance, while ablations confirm the contribution of latent mode control, data composition, reward shaping, and reasoning budget.
We believe \ours provides a useful perspective for future IQA research on latent-space reasoning, and may be extended to other perceptual rating tasks or combined with agentic visual tools for more controllable quality analysis.

\section*{Acknowledgment}

The authors appreciate the funding from the University of Bristol, and the UKRI MyWorld Strength in Places Programme (SIPF00006/1). We also acknowledge the use of resources provided by the Isambard-AI National AI Research Resource (AIRR). Isambard-AI is operated by the University of Bristol and is funded by the UK Government’s Department for Science, Innovation and Technology (DSIT) via UK Research and Innovation; and the Science and Technology Facilities Council [ST/AIRR/I-A-I/1023].

\bibliographystyle{plain}
\bibliography{main}

@String(CVPR  = {IEEE Conf. Comput. Vis. Pattern Recog.})

@String(ICCV  = {Int. Conf. Comput. Vis.})

@String(ECCV  = {Eur. Conf. Comput. Vis.})

@String(NeurIPS = {Adv. Neural Inform. Process. Syst.})

@String(ICLR  = {Int. Conf. Learn. Represent.})

@String(AAAI  = {AAAI})

@String(ICME  = {Int. Conf. Multimedia and Expo})

@String(TIP   = {IEEE Trans. Image Process.})

@String(TCSVT = {IEEE Trans. Circuit Syst. Video Technol.})

@String(CVPR  = {CVPR})

@String(ICCV  = {ICCV})

@String(ECCV  = {ECCV})

@String(NeurIPS = {NeurIPS})

@String(ICLR  = {ICLR})

@String(ICME  =	{ICME})

@String(TIP   = {IEEE TIP})

@String(TCSVT = {IEEE TCSVT})

@article{zhao2025reasoning,
  title={{Reasoning as Representation: Rethinking Visual Reinforcement Learning in Image Quality Assessment}},
  author={Zhao, Shijie and Zhang, Xuanyu and Li, Weiqi and Li, Junlin and Zhang, Li and Xue, Tianfan and Zhang, Jian},
  journal={arXiv preprint arXiv:2510.11369},
  year={2025}
}

@inproceedings{zhang2025beyond,
  title={{Beyond Text-Visual Attention: Exploiting Visual Cues for Effective Token Pruning in {VLMs}}},
  author={Zhang, Qizhe and Cheng, Aosong and Lu, Ming and Zhang, Renrui and Zhuo, Zhiyong and Cao, Jiajun and Guo, Shaobo and She, Qi and Zhang, Shanghang},
  booktitle={Proceedings of the IEEE/CVF International Conference on Computer Vision},
  pages={20857--20867},
  year={2025}
}

@article{cai2025diagnosing,
  title={{Diagnosing and Mitigating Modality Interference in Multimodal Large Language Models}},
  author={Cai, Rui and Li, Bangzheng and Wen, Xiaofei and Chen, Muhao and Zhao, Zhe},
  journal={arXiv preprint arXiv:2505.19616},
  year={2025}
}

@article{chen2026show,
  title={{Show, Don't Tell: Morphing Latent Reasoning into Image Generation}},
  author={Chen, Harold Haodong and Yin, Xinxiang and Shu, Wen-Jie and Zhang, Hongfei and Zhang, Zixin and Liao, Chenfei and Guo, Litao and Chen, Qifeng and Chen, Ying-Cong},
  journal={arXiv preprint arXiv:2602.02227},
  year={2026}
}

@article{wang2025monet,
  title={{Monet: Reasoning in Latent Visual Space Beyond Images and Language}},
  author={Wang, Qixun and Shi, Yang and Wang, Yifei and Zhang, Yuanxing and Wan, Pengfei and Gai, Kun and Ying, Xianghua and Wang, Yisen},
  journal={arXiv preprint arXiv:2511.21395},
  year={2025}
}

@article{mittal2012making,
  title   = {{Making a ``Completely Blind'' Image Quality Analyzer}},
  author  = {Mittal, Anish and Soundararajan, Rajiv and Bovik, Alan C.},
  journal = {IEEE Signal Processing Letters},
  volume  = {20},
  number  = {3},
  pages   = {209--212},
  year    = {2013}
}

@article{mittal2012no,
  title   = {{No-Reference Image Quality Assessment in the Spatial Domain}},
  author  = {Mittal, Anish and Moorthy, Anush Krishna and Bovik, Alan C.},
  journal = {IEEE Transactions on Image Processing (TIP)},
  volume  = {21},
  number  = {12},
  pages   = {4695--4708},
  year    = {2012}
}

@article{talebi2018nima,
  title   = {{NIMA}: {Neural Image Assessment}},
  author  = {Talebi, Hossein and Milanfar, Peyman},
  journal = {IEEE Transactions on Image Processing (TIP)},
  volume  = {27},
  pages   = {3998--4011},
  year    = {2018}
}

@inproceedings{ke2021musiq,
  author    = {Ke, Junjie and Wang, Qifei and Wang, Yilin and Milanfar, Peyman and Yang, Feng},
  title     = {{MUSIQ}: {Multi-Scale Image Quality Transformer}},
  booktitle = {Proceedings of the IEEE/CVF International Conference on Computer Vision (ICCV)},
  year      = {2021},
  pages     = {5148--5157}
}

@inproceedings{wang2023exploring,
  author    = {Wang, Jianyi and Chan, Kelvin C. K. and Loy, Chen Change},
  title     = {{Exploring CLIP for Assessing the Look and Feel of Images}},
  booktitle = {Proceedings of the AAAI Conference on Artificial Intelligence (AAAI)},
  year      = {2023}
}

@inproceedings{yang2022maniqa,
  author    = {Yang, Sidi and Wu, Tianhe and Shi, Shuwei and Lao, Shanshan and Gong, Yuan and Cao, Mingdeng and Wang, Jiahao and Yang, Yujiu},
  title     = {{MANIQA: Multi-Dimension Attention Network for No-Reference Image Quality Assessment}},
  booktitle = {Proceedings of the IEEE/CVF Conference on Computer Vision and Pattern Recognition (CVPR) Workshops},
  month     = {June},
  year      = {2022},
  pages     = {1191--1200}
}

@article{zhuadaptive,
  title   = {{Adaptive Image Quality Assessment via Teaching Large Multimodal Model to Compare}},
  author  = {Zhu, Hanwei and Wu, Haoning and Li, Yixuan and Zhang, Zicheng and Chen, Baoliang and Zhu, Lingyu and Fang, Yuming and Zhai, Guangtao and Lin, Weisi and Wang, Shiqi},
  journal = {arXiv preprint arXiv:2405.19298},
  year    = {2024}
}

@article{bai2025qwen2,
  title   = {{Qwen2.5-VL Technical Report}},
  author  = {Bai, Shuai and Chen, Keqin and Liu, Xuejing and Wang, Jialin and Ge, Wenbin and Song, Sibo and Dang, Kai and Wang, Peng and Wang, Shijie and Tang, Jun and Zhong, Humen and Zhu, Yuanzhi and Yang, Mingkun and Li, Zhaohai and Wan, Jianqiang and Wang, Pengfei and Ding, Wei and Fu, Zheren and Xu, Yiheng and Ye, Jiabo and Zhang, Xi and Xie, Tianbao and Cheng, Zesen and Zhang, Hang and Yang, Zhibo and Xu, Haiyang and Lin, Junyang and others},
  journal = {arXiv preprint arXiv:2502.13923},
  year    = {2025}
}

@inproceedings{wu2023QAlign,
  title     = {{Q-Align: Teaching LMMs for Visual Scoring via Discrete Text-Defined Levels}},
  author    = {Wu, Haoning and Zhang, Zicheng and Zhang, Weixia and Chen, Chaofeng and Liao, Liang and Li, Chunyi and Gao, Yixuan and Wang, Annan and Zhang, Erli and Sun, Wenxiu and Yan, Qiong and Min, Xiongkuo and Zhai, Guangtao and Lin, Weisi},
  booktitle = {Proceedings of the 41st International Conference on Machine Learning},
  series    = {Proceedings of Machine Learning Research},
  volume    = {235},
  pages     = {54015--54029},
  year      = {2024},
  publisher = {PMLR}
}

@inproceedings{you2025teaching,
  author    = {You, Zhiyuan and Cai, Xin and Gu, Jinjin and Xue, Tianfan and Dong, Chao},
  title     = {{Teaching Large Language Models to Regress Accurate Image Quality Scores Using Score Distribution}},
  booktitle = {Proceedings of the IEEE/CVF Conference on Computer Vision and Pattern Recognition (CVPR)},
  month     = {June},
  year      = {2025},
  pages     = {14483--14494}
}

@article{li2025qinsight,
  title={Q-insight: Understanding image quality via visual reinforcement learning},
  author={Li, Weiqi and Zhang, Xuanyu and Zhao, Shijie and Zhang, Yabin and Li, Junlin and Zhang, Li and Zhang, Jian},
  journal={arXiv preprint arXiv:2503.22679},
  year={2025}
}

@article{wu2025visualqualityr1,
  title   = {{VisualQuality-R1: Reasoning-Induced Image Quality Assessment via Reinforcement Learning to Rank}},
  author  = {Wu, Tianhe and Zou, Jian and Liang, Jie and Zhang, Lei and Ma, Kede},
  journal = {arXiv preprint arXiv:2505.14460},
  year    = {2025}
}

@article{chen2024expanding,
  title={{Expanding Performance Boundaries of Open-Source Multimodal Models with Model, Data, and Test-Time Scaling}},
  author={Chen, Zhe and Wang, Weiyun and Cao, Yue and Liu, Yangzhou and Gao, Zhangwei and Cui, Erfei and Zhu, Jinguo and Ye, Shenglong and Tian, Hao and Liu, Zhaoyang and others},
  journal={arXiv preprint arXiv:2412.05271},
  year={2024}
}

@inproceedings{loshchilov2019decoupled,
  title     = {{Decoupled Weight Decay Regularization}},
  author    = {Loshchilov, Ilya and Hutter, Frank},
  booktitle = {International Conference on Learning Representations (ICLR)},
  year      = {2019}
}

@article{hosu2020koniq,
  title={{KonIQ-10k}: An ecologically valid database for deep learning of blind image quality assessment},
  author={Hosu, Vlad and Lin, Hanhe and Sziranyi, Tamas and Saupe, Dietmar},
  journal={IEEE Transactions on Image Processing (TIP)},
  volume={29},
  pages={4041--4056},
  year={2020},
  publisher={IEEE}
}

@inproceedings{fang2020perceptual,
  title={{Perceptual Quality Assessment of Smartphone Photography}},
  author={Fang, Yuming and Zhu, Hanwei and Zeng, Yan and Ma, Kede and Wang, Zhou},
  booktitle={Proceedings of the IEEE/CVF Conference on Computer Vision and Pattern Recognition (CVPR)},
  pages={3677--3686},
  year={2020}
}

@article{ghadiyaram2015live,
  title={{LIVE In the Wild Image Quality Challenge Database}},
  author={Ghadiyaram, Deepti and Bovik, Alan C},
  journal={Online},
  year={2015}
}

@inproceedings{jinjin2020pipal,
  title={{PIPAL: a Large-Scale Image Quality Assessment Dataset for Perceptual Image Restoration}},
  author={GU, Jinjin and Cai, Haoming and Chen, Haoyu  and Ye, Xiaoxing and Jimmy S, Ren  and Dong, Chao},
  booktitle={Proceedings of the European Conference on Computer Vision (ECCV)},
  pages={633--651},
  year={2020},
  organization={Springer}
}

@article{li2023agiqa,
  title={{AGIQA-3K: An Open Database for AI-Generated Image Quality Assessment}},
  author={Li, Chunyi and Zhang, Zicheng and Wu, Haoning and Sun, Wei and Min, Xiongkuo and Liu, Xiaohong and Zhai, Guangtao and Lin, Weisi},
  journal={IEEE Transactions on Circuits and Systems for Video Technology (TCSVT)},
  volume={34},
  number={8},
  pages={6833--6846},
  year={2023},
  publisher={IEEE}
}

@inproceedings{lin2019kadid,
  title={{KADID-10k: A Large-scale Artificially Distorted IQA Database}},
  author={Lin, Hanhe and Hosu, Vlad and Saupe, Dietmar},
  booktitle={Proceedings of International Conference on Quality of Multimedia Experience (QoMEX)},
  pages={1--3},
  year={2019},
  organization={IEEE}
}

@article{larson2010most,
  title={Most apparent distortion: full-reference image quality assessment and the role of strategy},
  author={Larson, Eric C and Chandler, Damon M},
  journal={Journal of Electronic Imaging},
  volume={19},
  number={1},
  pages={011006--011006},
  year={2010},
  publisher={Society of Photo-Optical Instrumentation Engineers}
}

@article{deepfl-iqa,
title={{DeepFL-IQA: Weak Supervision for Deep IQA Feature Learning}},
author={Lin, Hanhe and Hosu, Vlad and Saupe, Dietmar},
journal={arXiv preprint arXiv:2001.08113},
year={2020}}

@article{shao2024visual,
  title={Visual cot: Unleashing chain-of-thought reasoning in multi-modal language models},
  author={Shao, Hao and Qian, Shengju and Xiao, Han and Song, Guanglu and Zong, Zhuofan and Wang, Letian and Liu, Yu and Li, Hongsheng},
  journal={arXiv preprint arXiv:2403.16999},
  volume={2},
  year={2024}
}

@book{moravec1988mindchildren,
  title     = {{Mind Children: The Future of Robot and Human Intelligence}},
  author    = {Moravec, Hans P.},
  year      = {1988},
  publisher = {Harvard University Press},
  address   = {Cambridge, MA}
}

@inproceedings{tian2025aigivc,
  title={{AI-generated Image Quality Assessment in Visual Communication}},
  author={Tian, Yu and Li, Yixuan and Chen, Baoliang and Zhu, Hanwei and Wang, Shiqi and Kwong, Sam},
  booktitle={Proceedings of the AAAI Conference on Artificial Intelligence},
  volume={39-7},
  pages={7392--7400},
  year={2025}
}

@article{goyal2023think,
  title={{Think before you speak: Training Language Models With Pause Tokens}},
  author={Goyal, Sachin and Ji, Ziwei and Rawat, Ankit Singh and Menon, Aditya Krishna and Kumar, Sanjiv and Nagarajan, Vaishnavh},
  journal={arXiv preprint arXiv:2310.02226},
  year={2023}
}

@article{hao2024training,
  title={{Training Large Language Models to Reason in a Continuous Latent Space}},
  author={Hao, Shibo and Sukhbaatar, Sainbayar and Su, DiJia and Li, Xian and Hu, Zhiting and Weston, Jason and Tian, Yuandong},
  journal={arXiv preprint arXiv:2412.06769},
  year={2024}
}

@inproceedings{shen2025codi,
  title={{CODI: Compressing Chain-of-Thought into Continuous Space via Self-Distillation}},
  author={Shen, Zhenyi and Yan, Hanqi and Zhang, Linhai and Hu, Zhanghao and Du, Yali and He, Yulan},
  booktitle={Proceedings of the 2025 Conference on Empirical Methods in Natural Language Processing},
  pages={677--693},
  year={2025}
}

@article{cheng2024compressed,
  title={{Compressed Chain of Thought: Efficient Reasoning Through Dense Representations}},
  author={Cheng, Jeffrey and Van Durme, Benjamin},
  journal={arXiv preprint arXiv:2412.13171},
  year={2024}
}

@article{yu2024rlaif,
  title={{RLAIF-V: Aligning MLLMs through Open-Source AI Feedback for Super GPT-4V Trustworthiness}},
  author={Yu, Tianyu and Zhang, Haoye and Yao, Yuan and Dang, Yunkai and Chen, Da and Lu, Xiaoman and Cui, Ganqu and He, Taiwen and Liu, Zhiyuan and Chua, Tat-Seng and others},
  journal={arXiv preprint arXiv:2405.17220},
  year={2024}
}

@inproceedings{zhao2025beyond,
  title={{Beyond Multimodal Hallucinations: Enhancing {LVLMs} through Hallucination-Aware Direct Preference Optimization}},
  author={Zhao, Zhiyuan and Wang, Bin and Ouyang, Linke and Dong, Xiaoyi and Wang, Jiaqi and He, Conghui},
  booktitle={2025 IEEE International Conference on Multimedia and Expo (ICME)},
  pages={1--6},
  year={2025},
  organization={IEEE}
}

@article{zhang2025r1vl,
  title={{R1-VL: Learning to Reason with Multimodal Large Language Models via Step-wise Group Relative Policy Optimization}},
  author={Zhang, Jingyi and Huang, Jiaxing and Yao, Huanjin and Liu, Shunyu and Zhang, Xikun and Lu, Shijian and Tao, Dacheng},
  journal={arXiv preprint arXiv:2503.12937},
  year={2025}
}

@article{sun2023aligning,
  title={{Aligning Large Multimodal Models with Factually Augmented RLHF}},
  author={Sun, Zhiqing and Shen, Sheng and Cao, Shengcao and Liu, Haotian and Li, Chunyuan and Shen, Yikang and Gan, Chuang and Gui, Liang-Yan and Wang, Yu-Xiong and Yang, Yiming and others},
  journal={arXiv preprint arXiv:2309.14525},
  year={2023}
}

@article{cai2025q,
  title={{Q-Ponder: A Unified Training Pipeline for Reasoning-based Visual Quality Assessment}},
  author={Cai, Zhuoxuan and Zhang, Jian and Yuan, Xinbin and Jiang, Pengtao and Chen, Wenxiang and Tang, Bowen and Yao, Lujian and Wang, Qiyuan and Chen, Jinwen and Li, Bo},
  journal={arXiv preprint arXiv:2506.05384},
  year={2025}
}

@article{li2025latent,
  title={Latent visual reasoning},
  author={Li, Bangzheng and Sun, Ximeng and Liu, Jiang and Wang, Ze and Wu, Jialian and Yu, Xiaodong and Chen, Hao and Barsoum, Emad and Chen, Muhao and Liu, Zicheng},
  journal={arXiv preprint arXiv:2509.24251},
  year={2025}
}

@article{li2025latentimplicitvisualreasoning,
  title={Latent Implicit Visual Reasoning},
  author={Li, Kelvin and Shang, Chuyi and Karlinsky, Leonid and Feris, Rogerio and Darrell, Trevor and Herzig, Roei},
  journal={arXiv preprint arXiv:2512.21218},
  year={2025}
}

@article{qin2025covt,
  title={Chain-of-visual-thought: Teaching {VLMs} to see and think better with continuous visual tokens},
  author={Qin, Yiming and Wei, Bomin and Ge, Jiaxin and Kallidromitis, Konstantinos and Fu, Stephanie and Darrell, Trevor and Wang, XuDong},
  journal={arXiv preprint arXiv:2511.19418},
  year={2025}
}

@article{liang2026zoomiqa,
  title={Zoom-{IQA}: Image Quality Assessment with Reliable Region-Aware Reasoning},
  author={Liang, Guoqiang and Wang, Jianyi and Wu, Zhonghua and Zhou, Shangchen},
  journal={arXiv preprint arXiv:2601.02918},
  year={2026}
}

@inproceedings{wei2022cot,
  title     = {{Chain-of-Thought Prompting Elicits Reasoning in Large Language Models}},
  author    = {Wei, Jason and Wang, Xuezhi and Schuurmans, Dale and Bosma, Maarten and Ichter, Brian and Xia, Fei and Chi, Ed H. and Le, Quoc V. and Zhou, Denny},
  booktitle = {Advances in Neural Information Processing Systems},
  volume    = {35},
  year      = {2022}
}

@inproceedings{wang2023selfconsistency,
  title     = {{Self-Consistency Improves Chain of Thought Reasoning in Language Models}},
  author    = {Wang, Xuezhi and Wei, Jason and Schuurmans, Dale and Le, Quoc V. and Chi, Ed H. and Narang, Sharan and Chowdhery, Aakanksha and Zhou, Denny},
  booktitle = {International Conference on Learning Representations (ICLR)},
  year      = {2023}
}

@inproceedings{zhou2023leasttomost,
  title     = {{Least-to-Most Prompting Enables Complex Reasoning in Large Language Models}},
  author    = {Zhou, Denny and Sch{\"a}rli, Nathanael and Hou, Le and Wei, Jason and Scales, Nathan and Wang, Xuezhi and Schuurmans, Dale and Cui, Claire and Bousquet, Olivier and Le, Quoc and Chi, Ed H.},
  booktitle = {International Conference on Learning Representations (ICLR)},
  year      = {2023}
}

@inproceedings{yao2023tot,
  title     = {{Tree of Thoughts: Deliberate Problem Solving with Large Language Models}},
  author    = {Yao, Shunyu and Yu, Dian and Zhao, Jeffrey and Shafran, Izhak and Griffiths, Thomas L. and Cao, Yuan and Narasimhan, Karthik},
  booktitle = {Advances in Neural Information Processing Systems},
  volume    = {36},
  year      = {2023}
}

@article{huang2025visionr1,
  title={Vision-r1: Incentivizing reasoning capability in multimodal large language models},
  author={Huang, Wenxuan and Jia, Bohan and Zhai, Zijie and Cao, Shaosheng and Ye, Zheyu and Zhao, Fei and Xu, Zhe and Hu, Yao and Lin, Shaohui},
  journal={arXiv preprint arXiv:2503.06749},
  year={2025}
}

@article{liu2025visual,
  title={{Visual-RFT: Visual Reinforcement Fine-Tuning}},
  author={Liu, Ziyu and Sun, Zeyi and Zang, Yuhang and Dong, Xiaoyi and Cao, Yuhang and Duan, Haodong and Lin, Dahua and Wang, Jiaqi},
  journal={arXiv preprint arXiv:2503.01785},
  year={2025}
}

@inproceedings{wu2024qinstruct,
  title={{Q-Instruct: Improving Low-level Visual Abilities for Multi-modality Foundation Models}},
  author={Wu, Haoning and Zhang, Zicheng and Zhang, Erli and Chen, Chaofeng and Liao, Liang and Wang, Annan and Xu, Kaixin and Li, Chunyi and Hou, Jingwen and Zhai, Guangtao and others},
  booktitle={Proceedings of the IEEE/CVF Conference on Computer Vision and Pattern Recognition (CVPR)},
  pages={25490--25500},
  year={2024}
}

@inproceedings{you2024DQA,
  title={{Depicting Beyond Scores: Advancing Image Quality Assessment through Multi-modal Language Models}},
  author={You, Zhiyuan and Li, Zheyuan and Gu, Jinjin and Yin, Zhenfei and Xue, Tianfan and Dong, Chao},
  booktitle={Proceedings of the European Conference on Computer Vision (ECCV)},
  pages={259--276},
  year={2024},
  organization={Springer}
}

@inproceedings{wu2024coinstruct,
  title     = {{Towards Open-ended Visual Quality Comparison}},
  author    = {Wu, Haoning and Zhu, Hanwei and Zhang, Zicheng and Zhang, Erli and Chen, Chaofeng and Liao, Liang and Li, Chunyi and Wang, Annan and Sun, Wenxiu and Yan, Qiong and Liu, Xiaohong and Zhai, Guangtao and Wang, Shiqi and Lin, Weisi},
  booktitle = {European Conference on Computer Vision (ECCV)},
  year      = {2024}
}

@article{zhang2025vqinsight,
  title={{VQ-Insight: Teaching VLMs for AI-Generated Video Quality Understanding via Progressive Visual Reinforcement Learning}},
  author={Zhang, Xuanyu and Li, Weiqi and Zhao, Shijie and Li, Junlin and Zhang, Li and Zhang, Jian},
  journal={Proceedings of the AAAI Conference on Artificial Intelligence (AAAI)},
  year={2026}
}

@article{yang2025mirage,
  title={Machine mental imagery: Empower multimodal reasoning with latent visual tokens},
  author={Yang, Zeyuan and Yu, Xueyang and Chen, Delin and Shen, Maohao and Gan, Chuang},
  journal={arXiv preprint arXiv:2506.17218},
  year={2025}
}

@article{chen2025reasoning,
  title={Reasoning in the dark: Interleaved vision-text reasoning in latent space},
  author={Chen, Chao and Ma, Zhixin and Li, Yongqi and Hu, Yupeng and Wei, Yinwei and Li, Wenjie and Nie, Liqiang},
  journal={arXiv preprint arXiv:2510.12603},
  year={2025}
}

@article{ma2025cocova,
  title={Cocova: Chain of continuous vision-language thought for latent space reasoning},
  author={Ma, Jizheng and Zhou, Xiaofei and Song, Yanlong and Yan, Han},
  journal={arXiv e-prints},
  pages={arXiv--2511},
  year={2025}
}

@article{liu2023visual,
  title={{Visual Instruction Tuning}},
  author={Liu, Haotian and Li, Chunyuan and Wu, Qingyang and Lee, Yong Jae},
  journal={Proceedings of the Advances in Neural Information Processing Systems (NeurIPS)},
  volume={36},
  pages={34892--34916},
  year={2023}
}

@article{you2024DQAW,
  title={{Descriptive Image Quality Assessment in the Wild}},
  author={You, Zhiyuan and Gu, Jinjin and Li, Zheyuan and Cai, Xin and Zhu, Kaiwen and Dong, Chao and Xue, Tianfan},
  journal={arXiv preprint arXiv:2405.18842},
  year={2024}
}

@article{bosse2017deep,
  title={{Deep neural networks for no-reference and full-reference image quality assessment}},
  author={Bosse, Sebastian and Maniry, Dominique and M{\"u}ller, Klaus-Robert and Wiegand, Thomas and Samek, Wojciech},
  journal={IEEE Transactions on Image Processing (TIP)},
  volume={27},
  number={1},
  pages={206--219},
  year={2017},
  publisher={IEEE}
}

@inproceedings{zhu2020metaiqa,
  title={{MetaIQA: Deep Meta-learning for No-Reference Image Quality Assessment}},
  author={Zhu, Hancheng and Li, Leida and Wu, Jinjian and Dong, Weisheng and Shi, Guangming},
  booktitle={Proceedings of the IEEE/CVF Conference on Computer Vision and Pattern Recognition (CVPR)},
  pages={14143--14152},
  year={2020}
}

@article{chen2021no,
  title={{No-Reference Screen Content Image Quality Assessment With Unsupervised Domain Adaptation}},
  author={Chen, Baoliang and Li, Haoliang and Fan, Hongfei and Wang, Shiqi},
  journal={IEEE Transactions on Image Processing (TIP)},
  volume={30},
  pages={5463--5476},
  year={2021},
  publisher={IEEE}
}

@inproceedings{chen2025toward,
  title={{Toward Generalized Image Quality Assessment: Relaxing the Perfect Reference Quality Assumption}},
  author={Chen, Du and Wu, Tianhe and Ma, Kede and Zhang, Lei},
  booktitle={Proceedings of the IEEE/CVF Conference on Computer Vision and Pattern Recognition (CVPR)},
  pages={12742--12752},
  year={2025}
}

@inproceedings{wang2021troubleshooting,
  title={{Troubleshooting Blind Image Quality Models in the Wild}},
  author={Wang, Zhihua and Wang, Haotao and Chen, Tianlong and Wang, Zhangyang and Ma, Kede},
  booktitle={Proceedings of the IEEE/CVF Conference on Computer Vision and Pattern Recognition (CVPR)},
  pages={16256--16265},
  year={2021}
}

@article{zhang2022continual,
  title={{Continual Learning for Blind Image Quality Assessment}},
  author={Zhang, Weixia and Li, Dingquan and Ma, Chao and Zhai, Guangtao and Yang, Xiaokang and Ma, Kede},
  journal={IEEE Transactions on Pattern Analysis and Machine Intelligence},
  volume={45},
  number={3},
  pages={2864--2878},
  year={2022},
  publisher={IEEE}
}

@article{shao2024deepseekmath,
  title={Deepseekmath: Pushing the limits of mathematical reasoning in open language models},
  author={Shao, Zhihong and Wang, Peiyi and Zhu, Qihao and Xu, Runxin and Song, Junxiao and Bi, Xiao and Zhang, Haowei and Zhang, Mingchuan and Li, YK and Wu, Yang and others},
  journal={arXiv preprint arXiv:2402.03300},
  year={2024}
}

@article{guo2025deepseek,
  title={Deepseek-r1: Incentivizing reasoning capability in llms via reinforcement learning},
  author={Guo, Daya and Yang, Dejian and Zhang, Haowei and Song, Junxiao and Wang, Peiyi and Zhu, Qihao and Xu, Runxin and Zhang, Ruoyu and Ma, Shirong and Bi, Xiao and others},
  journal={arXiv preprint arXiv:2501.12948},
  year={2025}
}

@article{van2008visualizing,
  title={Visualizing data using {t-SNE}.},
  author={Van der Maaten, Laurens and Hinton, Geoffrey},
  journal={Journal of machine learning research},
  volume={9},
  number={11},
  year={2008}
}

@inproceedings{xu2018attngan,
  title={Attngan: Fine-grained text to image generation with attentional generative adversarial networks},
  author={Xu, Tao and Zhang, Pengchuan and Huang, Qiuyuan and Zhang, Han and Gan, Zhe and Huang, Xiaolei and He, Xiaodong},
  booktitle={Proceedings of the IEEE conference on computer vision and pattern recognition},
  pages={1316--1324},
  year={2018}
}

@article{ramesh2022hierarchical,
  title={Hierarchical text-conditional image generation with clip latents},
  author={Ramesh, Aditya and Dhariwal, Prafulla and Nichol, Alex and Chu, Casey and Chen, Mark},
  journal={arXiv preprint arXiv:2204.06125},
  year={2022}
}

@inproceedings{nichol2021glide,
  title={{GLIDE}: Towards Photorealistic Image Generation and Editing with Text-Guided Diffusion Models},
  author={Nichol, Alexander Quinn and Dhariwal, Prafulla and Ramesh, Aditya and Shyam, Pranav and Mishkin, Pamela and Mcgrew, Bob and Sutskever, Ilya and Chen, Mark},
  booktitle={International Conference on Machine Learning},
  pages={16784--16804},
  year={2022},
  organization={PMLR}
}

@misc{Midjourney,
   author = {Holz, David},
   title = {Midjourney},
   howpublished = {\url{https://www.midjourney.com/}},
   year = {2023}
}

@inproceedings{rombach2022high,
  title={High-resolution image synthesis with latent diffusion models},
  author={Rombach, Robin and Blattmann, Andreas and Lorenz, Dominik and Esser, Patrick and Ommer, Bj{\"o}rn},
  booktitle={Proceedings of the IEEE/CVF Conference on Computer Vision and Pattern Recognition},
  pages={10684--10695},
  year={2022}
}

@article{rombach2022text,
  title={Text-guided synthesis of artistic images with retrieval-augmented diffusion models},
  author={Rombach, Robin and Blattmann, Andreas and Ommer, Bj{\"o}rn},
  journal={arXiv preprint arXiv:2207.13038},
  year={2022}
}

@article{thomee2016yfcc100m,
  title={{YFCC100M}: The New Data in Multimedia Research},
  author={Thomee, Bart and Shamma, David A and Friedland, Gerald and Elizalde, Benjamin and Ni, Karl and Poland, Douglas and Borth, Damian and Li, Li-Jia},
  journal={Communications of the ACM},
  volume={59},
  number={2},
  pages={64--73},
  year={2016},
  publisher={ACM New York, NY, USA}
}

@incollection{mcintosh2024isambard,
  title={{Isambard-AI}: a leadership-class supercomputer optimised specifically for artificial intelligence},
  author={McIntosh-Smith, Simon and Alam, Sadaf and Woods, Christopher},
  booktitle={Proceedings of the Cray User Group},
  pages={44--54},
  year={2024}
}

\clearpage
\appendix

\setcounter{section}{0}
\renewcommand\thesection{\Alph{section}}
\setcounter{table}{0}
\renewcommand\thetable{\Alph{table}}
\makeatletter
  \renewcommand\theHtable{\Alph{table}}   
\makeatother
\setcounter{figure}{0}
\renewcommand\thefigure{\Alph{figure}}
\makeatletter
  \renewcommand\theHfigure{\Alph{figure}}
\makeatother

\section{More Details on Image Quality Assessment Datasets}\label{app:dataset}
The details of the utilized Image Quality Assessment (IQA) datasets are summarized as follows.
\begin{itemize}
\item  \textbf{CSIQ~\cite{larson2010most}:} The CSIQ database contains $30$ reference images and $866$ distorted images with $6$ distortions of JPEG compression, JPEG2000 compression, Gaussian blur, Gaussian white noise, Gaussian pink noise, and contrast change. The subjective testing method is the single-stimulus absolute category rating. The DMOSs span from $0$ to $1$.
\item \textbf{KADID-10k~\cite{lin2019kadid}:} The KADID-10k database contains $81$ reference images and $10,125$ distorted images by adding $25$ distortion types with $5$ distortion levels, such as blur, color distortions, noise, spatial distortions, etc. The subjective testing method is the double-stimulus continuous quality rating by crowdsourcing. The DMOSs range from $1$ to $5$.
\item \textbf{KonIQ-10k~\cite{hosu2020koniq}:} The KonIQ-10K database consists of $10,073$ images with abundant realistic distortions. Those are selected from the YFCC100M database~\cite{thomee2016yfcc100m}. Single-stimulus absolute category rating is the method of subjective testing. The MOSs range from $1$ to $5$.
\item \textbf{SPAQ~\cite{fang2020perceptual}:} The SPAQ database consists of $11,125$ in-the-wild pictures taken by $66$ smartphones. Each picture is annotated with quality, attributes, and scene categories using the single-stimulus methodology. The MOSs range from $0$ to $100$.
\item \textbf{AGIQA-3K~\cite{li2023agiqa}:} The AGIQA-3K consists of $2,982$ AI-generated images derived from $6$ advanced text-to-image generation models, which include AttnGAN~\cite{xu2018attngan}, DALLE2~\cite{ramesh2022hierarchical}, GLIDE~\cite{nichol2021glide}, Midjourney~\cite{Midjourney}, Stable Diffusion~\cite{rombach2022high}, and Stable Diffusion XL~\cite{rombach2022text}. The single-stimulus continuous quality rating is used to collect human opinions.  The MOSs range from $0$ to $5$.
\item \textbf{LIVE-Wild~\cite{ghadiyaram2015live}:} The LIVE In the Wild (LIVE Challenge) database contains $1,162$ authentically distorted images captured by diverse mobile devices in real-world conditions. The subjective testing method is large-scale crowdsourced continuous quality rating. The MOSs range from $0$ to $100$.
\item \textbf{PIPAL~\cite{jinjin2020pipal}:} The PIPAL dataset contains $250$ high-quality reference patches ($288\times288$) sampled from DIV2K/Flickr2K and $29$K distorted images generated from $40$ distortion/processing types (including traditional distortions, restoration/SR outputs, and GAN-based results), with $116$ distorted variants per reference. Subjective scores are collected via pairwise comparisons and aggregated using an Elo rating system. The MOSs are Elo-based scores (typically on the order of $\sim 1000$--$1800$).
\end{itemize}

\begin{figure}[!t]
    \centering
    \includegraphics[width=1\linewidth]{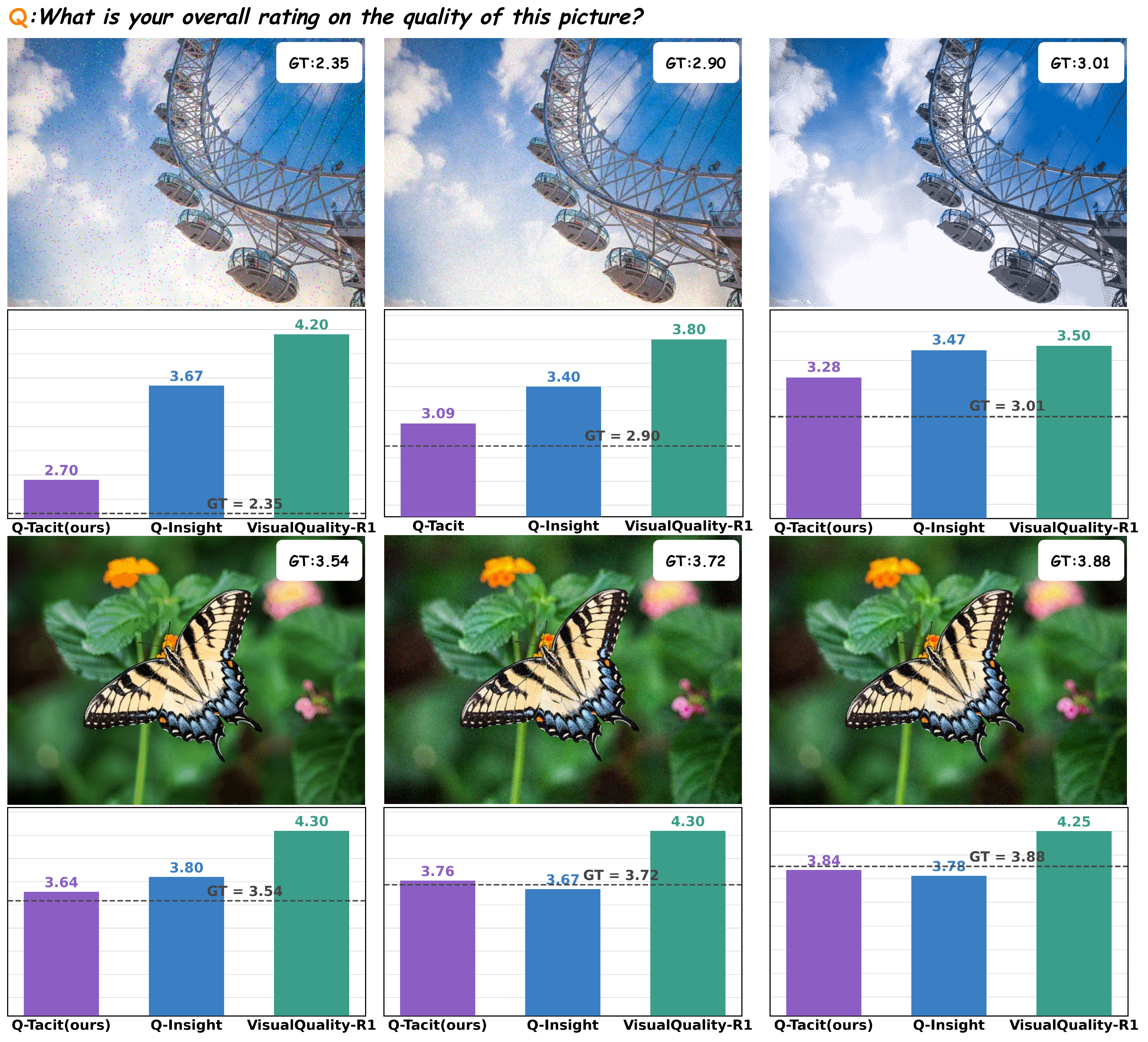}
    \caption{\textbf{Qualitative score prediction examples from KADID-10k~\cite{lin2019kadid}.}
For each example, the normalized MOS as ground-truth score (GT) is shown above the image, and the bar plot compares predicted scores from Q-Tacit, Q-Insight~\cite{li2025qinsight}, and VisualQuality-R1~\cite{wu2025visualqualityr1}.
The dashed line indicates GT for easier visual comparison. }
\label{fig:appendix_cases}
\end{figure}

\section{Additional Qualitative Results.}

The main results report aggregate correlation metrics across multiple benchmarks, while \cref{fig:appendix_cases} provides additional qualitative comparisons.
Across diverse content, Q-Tacit, Q-Insight~\cite{li2025qinsight}, and VisualQuality-R1~\cite{wu2025visualqualityr1} are prompted with the default rating query. Within each row, the three images are derived from the same source image but processed with different degradation settings (from KADID-10k~\cite{lin2019kadid}), resulting in a graded quality change with corresponding normalized MOS annotations.
This controlled setup enables a direct assessment of relative score behavior: Q-Tacit’s predictions track the MOS trend more closely within each row, whereas text-space baselines (Q-Insight and VisualQuality-R1) exhibit relative bias, over- or underestimating particular variants. Q-Tacit more consistently preserves the MOS ordering among subtly different variants in these examples.
These qualitative observations complement the main quantitative results by illustrating how the latent space supports consistent score ordering and stable per-image predictions.

\section{Model Complexity Comparison.}

\cref{tab:efficiency} compares the computational footprint of representative reasoning-based IQA methods under a unified setting (input resolution $512\times384\times3$).
By leveraging a compact latent reasoning segment, Q-Tacit achieves a favorable trade-off: it significantly reduces inference time relative to text-space reasoning baselines and lowers TFLOPs compared to Q-Insight/VisualQuality-R1, demonstrating the improved efficiency of compact tokens.

\begin{table}[t]
\centering
\scriptsize
\renewcommand{\arraystretch}{1.05}
\caption{Averaged model complexity comparison using $512\times 384\times 3$ images as input tested on single GH200 GPU.}
\resizebox{0.9\textwidth}{!}{
\begin{tabular}{l|c|c|c|c}
\toprule
\multicolumn{1}{l|}{Method} & \#Parameters & \begin{tabular}[c]{@{}c@{}}Inference \\ Time\end{tabular} $\downarrow$ & \begin{tabular}[c]{@{}c@{}} Inference\\ Memory\end{tabular} & \#TFLOPs $\downarrow$ \\ \hline
Q-Insight~\cite{li2025qinsight}  & 8.29 B      & 2.22 s      & 15.6 G       & 8.13   \\
VisualQuality-R1~\cite{wu2025visualqualityr1}            & 8.29 B      & 2.04 s      & 15.6 G       & 7.49   \\
\textbf{Q-Tacit(Ours)}            & 8.29 B      & 0.83 s      & 15.6 G       & 6.62   \\
\bottomrule
\end{tabular}
}
\label{tab:efficiency}
\end{table}

\end{document}